\documentclass[10pt,twocolumn,letterpaper]{article}

\usepackage{cvpr}
\usepackage{times}
\usepackage{epsfig}
\usepackage{graphicx}
\usepackage{amsmath}
\usepackage{amssymb}
\usepackage{kotex}
\usepackage{graphicx}
\usepackage{caption}
\usepackage{amsmath}
\usepackage{subcaption}
\usepackage{setspace}%
\usepackage{multirow}
\usepackage[table,xcdraw]{xcolor}
\usepackage{cprotect}

\usepackage[breaklinks=true,bookmarks=false]{hyperref}

\cvprfinalcopy 

\def\cvprPaperID{****} 

\begin{document}

\title{Small Object Detection using Context and Attention}

\author{Jeong-Seon Lim\\
{\tt\small University of Science and Technology}\\
{\tt\small jungsun0427@etri.re.kr}
\and
Marcella Astrid\\
{\tt\small University of Science and Technology}\\
{\tt\small marcella.astrid@ust.ac.kr}
\and
Hyun-Jin Yoon\\
{\tt\small Electronics and Telecommunications Research Institute}\\
{\tt\small hjyoon73@etri.re.kr}
\and
Seung-Ik Lee\\
{\tt\small Electronics and Telecommunications Research Institute}\\
{\tt\small the{\_}silee@etri.re.kr}
}

\maketitle

\begin{abstract}
   There are many limitations applying object detection algorithm on various environments.
   Especially detecting small objects is still challenging because they have low-resolution and limited information. 
   We propose an object detection method using context for improving accuracy of detecting small objects.
   The proposed method uses additional features from different layers as context by concatenating multi-scale features. 
   We also propose object detection with attention mechanism which can focus on the object in image, and it can include contextual information from target layer.
    Experimental results shows that proposed method also has higher accuracy than conventional SSD on detecting small objects.
   Also, for 300$\times$300 input, we achieved 78.1\verb|%| Mean Average Precision (mAP) on the PASCAL VOC2007 test set.

\end{abstract}

\section{Introduction}
\begin{figure}
	\centering
	\includegraphics[width=0.45\textwidth]{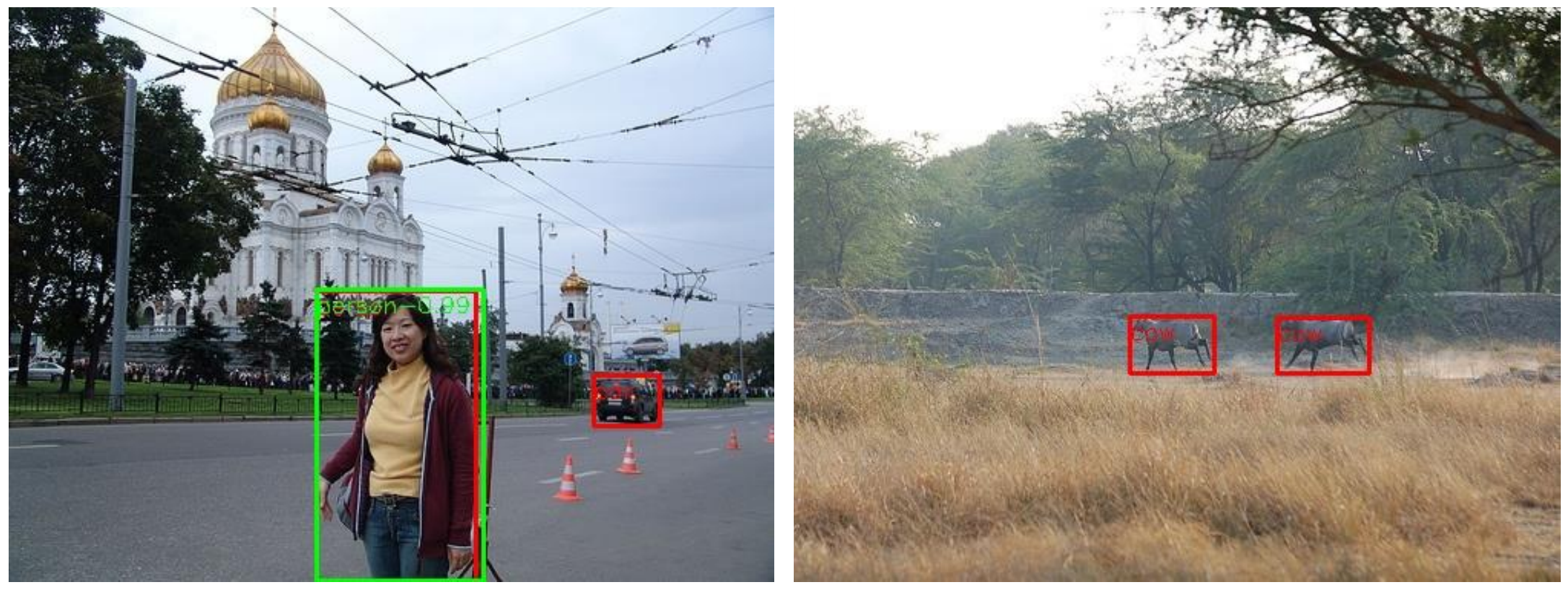}
	\caption{Failure cases of SSD in detecting small objects}
	\label{fig:1}
\end{figure}

Object detection is one of key topics in computer vision which th goals are finding bounding box of objects and their classification given an image.
In recent years, there has been huge improvements in accuracy and speed with the lead of deep learning technology: Faster R-CNN \cite{ren2015faster} achieved 73.2\verb|%| mAP, YOLOv2 \cite{redmon2017yolo9000} achieved 76.8\verb|%| mAP, SSD \cite{liu2016ssd} achieved 77.5\verb|%| mAP. 
However, there still remains important challenges in detecting small objects. For example, SSD can only achieved 20.7\verb|%| mAP on small objects targets.
Figure \ref{fig:1} shows the failure cases when SSD cannot detect the small objects. There are still a lot of room to improve in small object detection.


\begin{figure}
	\centering
	\includegraphics[width=0.45\textwidth]{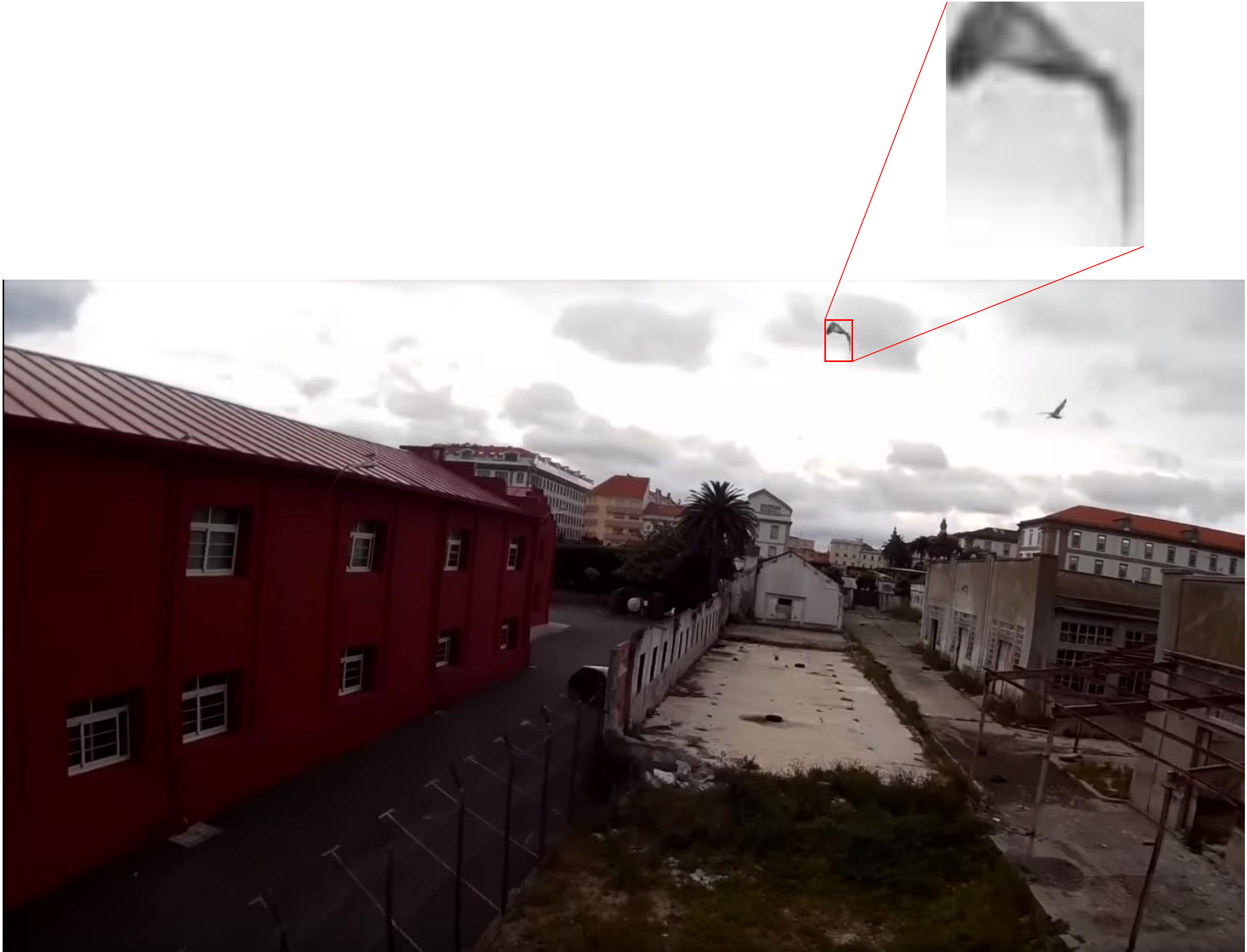}
	\caption{Context of small object is necessary to recognize \textit{bird} in this picture}
	\label{fig:2}
\end{figure}

Small object detection is difficult because of low-resolution and limited pixels.
For example, by looking only at the object on Figure \ref{fig:2}, it is even difficult for human to recognize the objects.
However, the object can be recognized as \textit{bird} by considering the context that it is located at sky.
Therefore, we believe that the key to solve this problem depends on how we can include context as extra information to help detecting small objects.


In this paper, we propose to use context information object for tackling the challenging problem of detecting small objects. 
First, to provide enough information on small objects, we extract context information from surrounded pixels of small objects by utilizing more abstract features from higher layers for the context of an object.
By concatenating the features of an small object and the features of the context, we augment the information for small objects so that the detector can detect the objects better.
Second, to focus on the small object, we use an attention mechanism in the early layer.
This is also help to reduce unnecessary shallow features information from background.
We select Single Shot Multibox Detector (SSD) \cite{liu2016ssd} for our baseline in our experiments. However, the idea can be generalize to other networks.
In order to evaluate the performance of the proposed model, we train our model to PASCAL VOC2007 and VOC2012 \cite{everingham2010pascal}, and comparison with baseline and state-of-the-art methods on VOC2007 will be given.

\section{Related Works}
\textbf{Object detection with deep learning} 
The advancement of deep learning technology has been improving the accuracy of object detection greatly. The first try for object detection with deep learning was R-CNN  \cite{girshick2014rich}.
R-CNN uses Convolutional Neural Network(CNN) on region proposals generated by using selective search \cite{uijlings2013selective}.
It is, however, too slow for real-time applications since each proposed region goes through CNNs sequentially.
Fast R-CNN\cite {girshick2015fast} is faster than R-CNN because it performs feature extraction stage only once for all the region proposals.
But those two works still use separate stage for region proposals, which becomes the main tackling point by Faster R-CNN \cite{ren2015faster} combines the region proposal phase and classification phase into one model such that it allows so-called end-to-end learning.
Object detection technologies has even been accelerated by YOLO \cite{redmon2016you} and SSD \cite{liu2016ssd} showing high performance enough for real-time object detection.
However, they are still not showing good performance for small objects.

\textbf{Small object detection}
Recently, several ideas has been proposed for detecting small object \cite{liu2016ssd, fu2017dssd, jeong2017enhancement, li2017perceptual}.
Liu et al \cite{liu2016ssd} augmented small object data by reducing the size of large objects for overcoming the not-enough-data problem.
Besides the approach for data augmentation, there has been some efforts for augmenting the required information without augmenting dataset perse.
DSSD \cite{fu2017dssd} applies deconvolution technique on all the feature maps of SSD to obtain scaled-up feature maps.
However, it has the limitation of increased model complexity and slow down an speed due to applying deconvolution module to all feature maps.
R-SSD \cite{jeong2017enhancement} combines features of different scales through pooling and deconvolution and obtained improved accuracy and speed compared to DSSD.
Li et al \cite{li2017perceptual} uses Generative Adversarial Network(GAN) \cite{goodfellow2014generative} to generate high-resolution features using low-resolution features as input to GAN.

\textbf{Visual attention network}
Attention mechanism in deep learning can be broadly understood as focusing on part of input for solving specific task rather than seeing the entire input.
Thus, attention mechanism is quite similar to what humans do when we see or hear something, 
Xu et al \cite{xu2015show} uses visual attention to generate image captions.
In order to generate caption corresponding to images, they used Long Short-Term Memory(LSTM) and the LSTM takes a relevant part of a given image.
Sharm et al \cite{sharma2015action} applied attention mechanism to recognize actions in video.
Wang et al \cite{wang2017residual} improved classification performance on ImageNet dataset by stacking residual attention modules. 


\begin{figure}
	\begin{subfigure}{8cm}
		\centering
		\includegraphics[width=\linewidth]{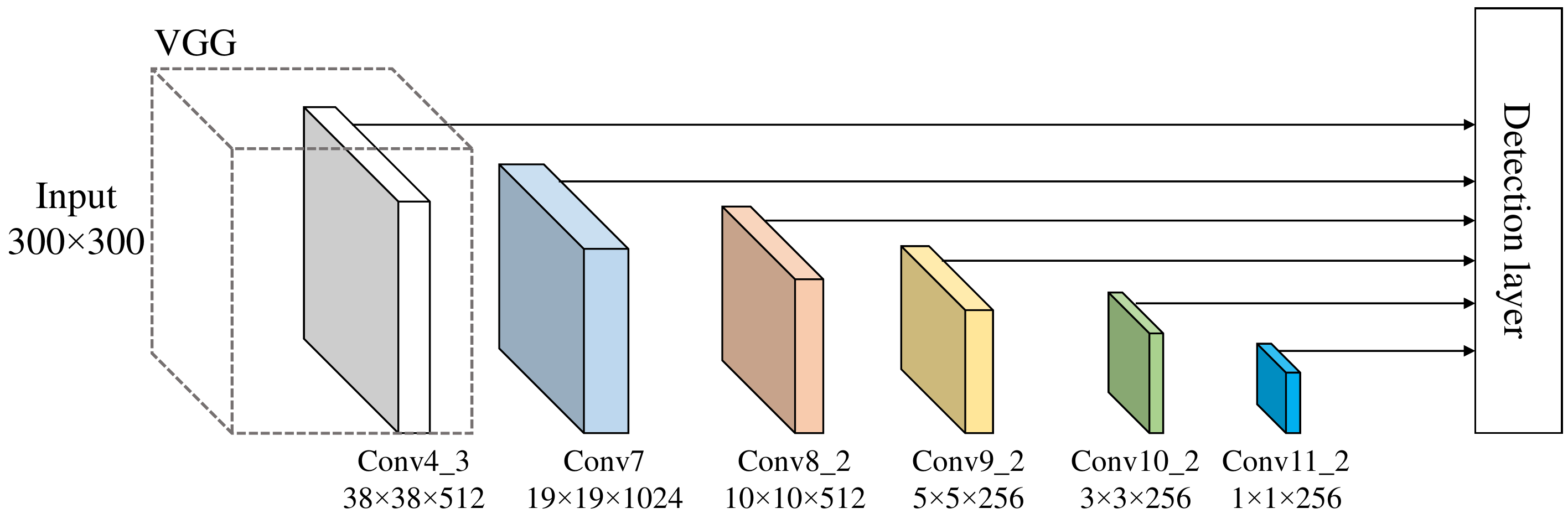}
		\caption{Conventional SSD with input 300$\times$300}
		\label{fig:SSD}
	\end{subfigure}
	
	\begin{subfigure}{8cm}
		\centering
		\includegraphics[width=\linewidth]{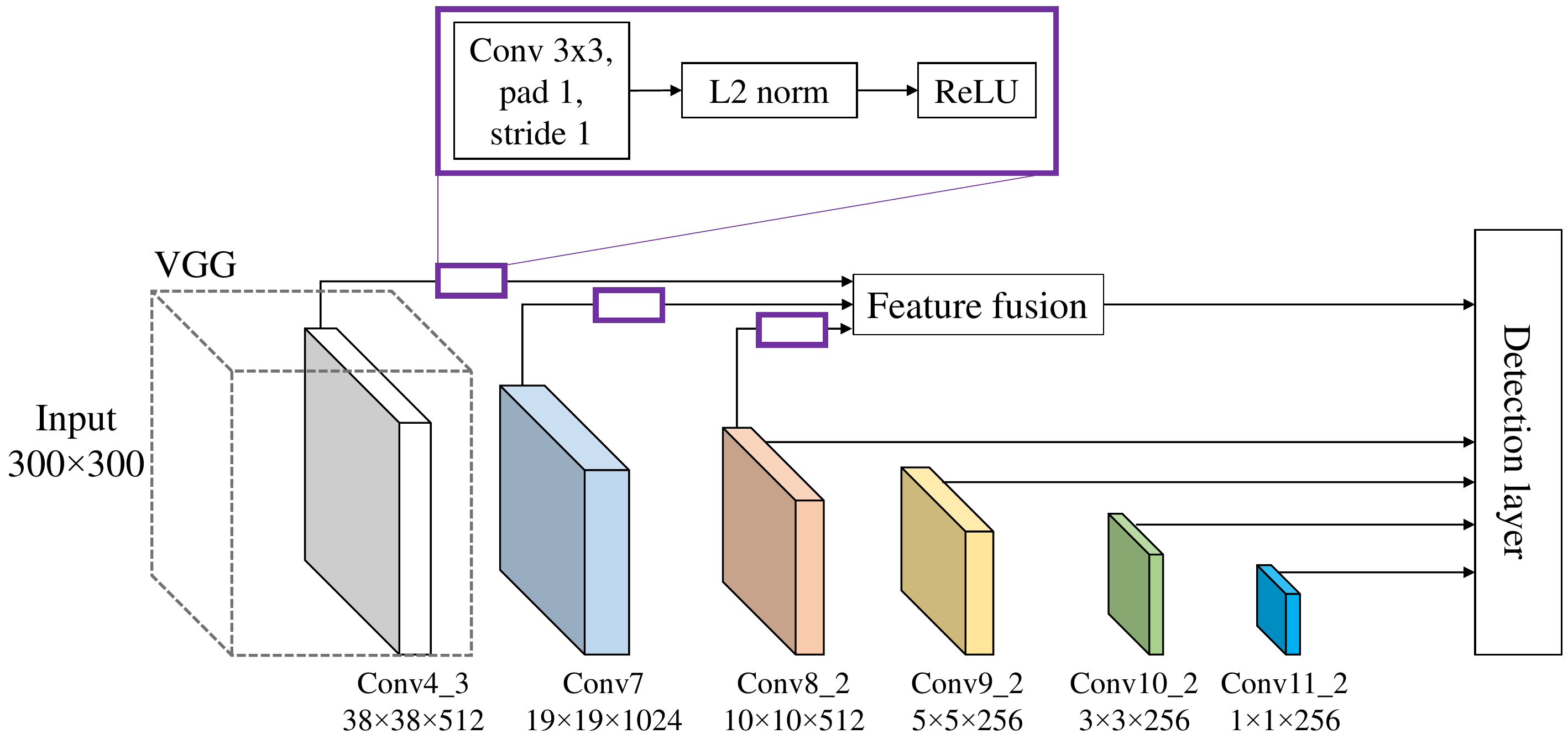}
		\caption{SSD with feature fusion (F-SSD)}
		\label{fig:F-SSD}
	\end{subfigure}
	
	\begin{subfigure}{8cm}
		\centering
		\includegraphics[width=\linewidth]{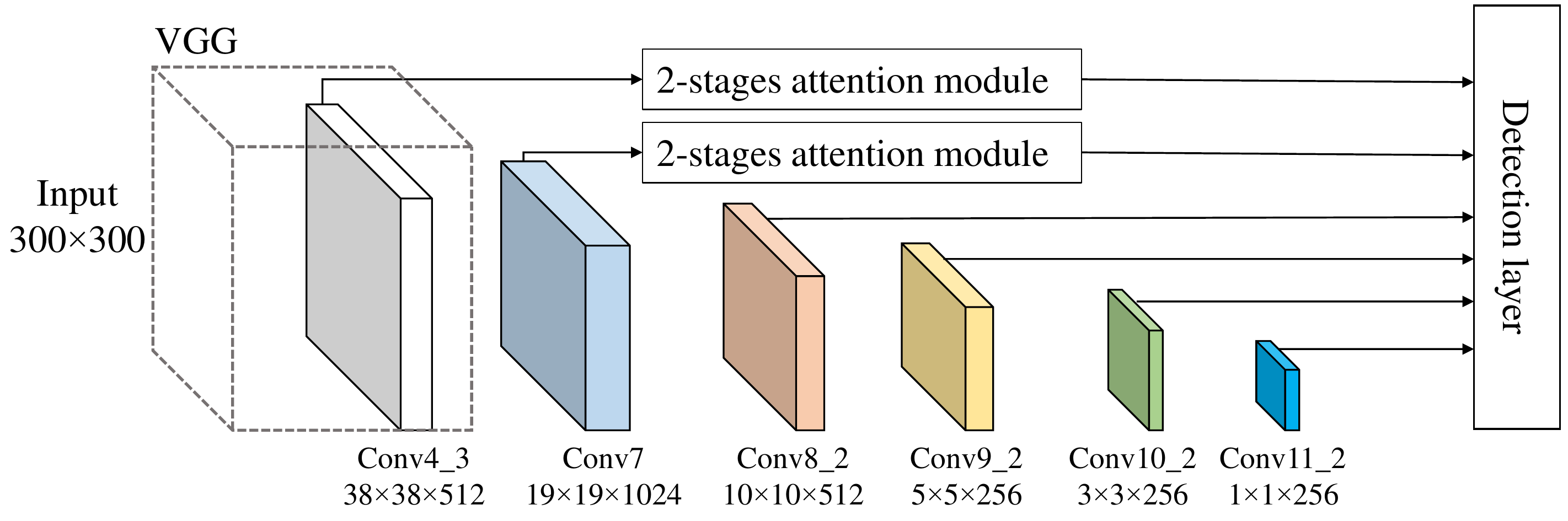}
		\caption{SSD with attention module (A-SSD)}
		\label{fig:A-SSD}
	\end{subfigure}

	\begin{subfigure}{8cm}
	\centering
	\includegraphics[width=\linewidth]{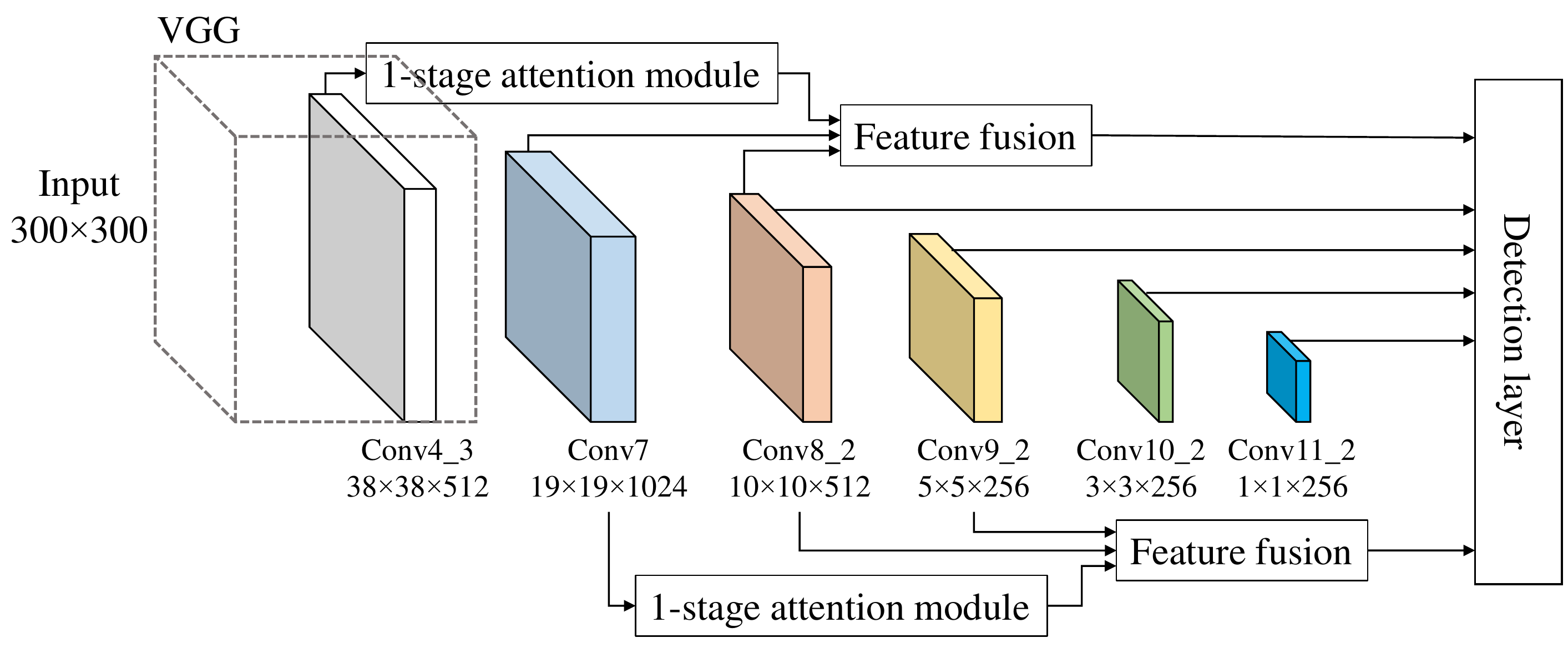}
	\caption{SSD with feature fusion + attention module (FA-SSD)}
	\label{fig:FA-SSD}
	\end{subfigure}

	\caption{Architectures of SSD and our approaches with VGG backbone.}\label{fig:architectures}
\end{figure}


\section{Method}
This section will discuss the baseline SSD, then followed by the components we propose to improve small object detection capability. First, SSD with feature fusion to get the context information, named F-SSD. Second, SSD with attention module to give the network capability to focus on important parts, named A-SSD. Third, we combine both feature fusion and attention module, named FA-SSD.
 
\subsection{Single Shot Multibox Detector (SSD)} \label{SSD}
In this section, we review Single Shot Multibox Detector (SSD) \cite{liu2016ssd}, which we are going to improve the capability on detecting small object. Like YOLO \cite{redmon2016you}, it is a one-stage detector which goal is to improve the speed, while also improving the detection in different scales by processing different level of feature maps, as seen in Fig. \ref{fig:SSD}. The idea is utilizing the higher resolution of early feature maps to detect smaller objects while the deeper feature which has lower resolution for the larger object detection. 

It is based on VGG16 \cite{simonyan2014very} backbone with additional layers to create different resolution of feature maps, as seen in Fig. \ref{fig:SSD}. From each of the features, with one additional convolution layer to match the output channels, the network predicts the output that consists both the bounding box regression and object classification.

However, the performance on small objects is still low, 20.7\verb|%| on VOC 2007, hence there are still many room for improvement. We believe there are two main reasons. First, the lack of context information to detect small object. On top of that, the features for small object detection are taken from shallow features which lack of semantic information. Our goal is to improve the SSD by adding feature fusion to solve the two problems. In addition, to improve more, we add attention module to make the network focuses only on the important part.
 
 \begin{figure}
 	\centering
 	\includegraphics[width=\linewidth]{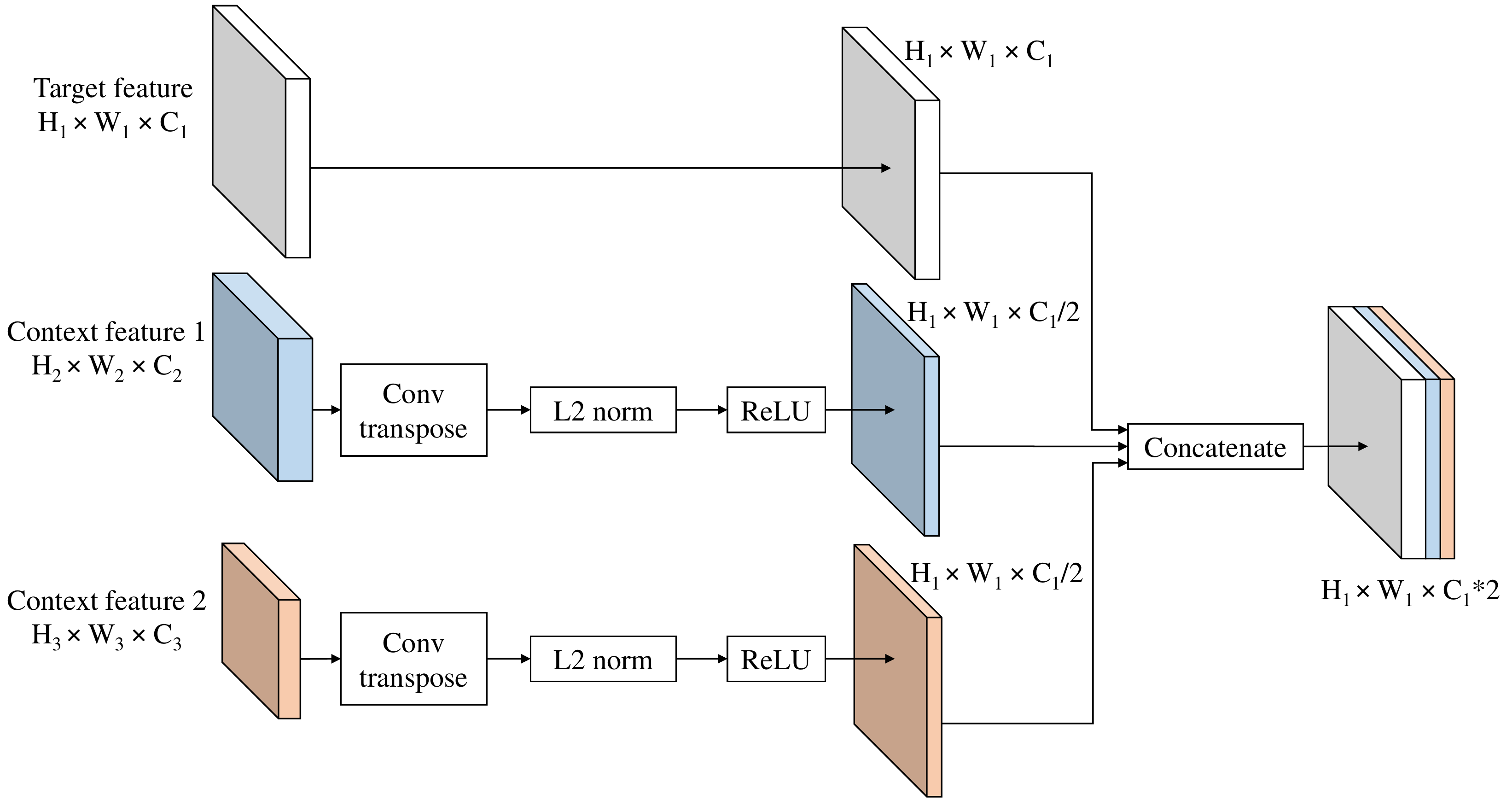}
 	\caption{Proposed feature fusion method}\label{fig:feature_fusion}
 \end{figure}
 
 
 \begin{figure}
 	\begin{subfigure}{8cm}
 		\centering
 		\includegraphics[width=1\linewidth]{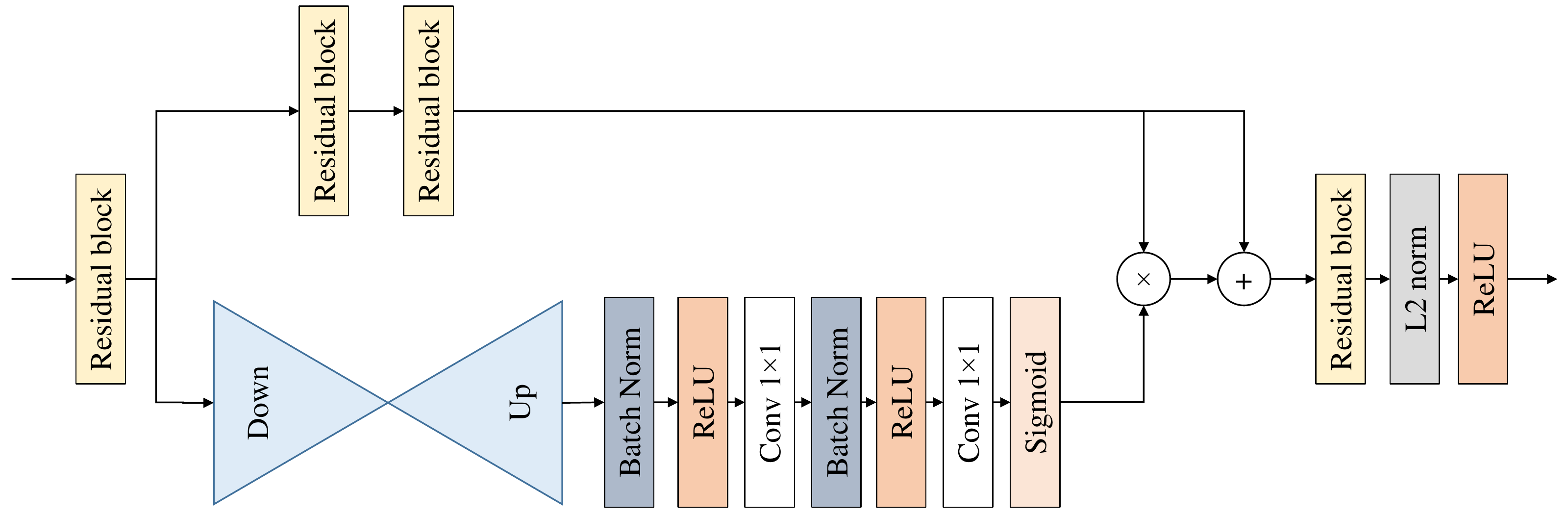}
 		\caption{Residual attention module \cite{wang2017residual}.}
 		\label{fig:residualattention}
 	\end{subfigure}
 	\begin{subfigure}{8cm}
 		\centering
 		\includegraphics[width=1\linewidth]{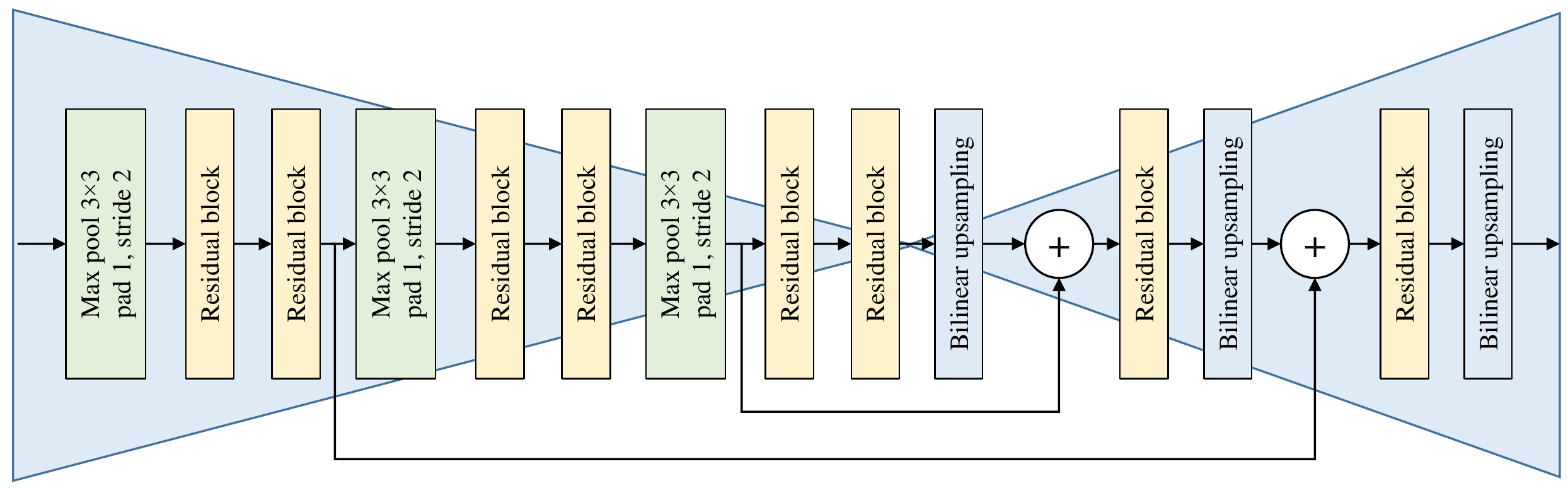}
 		\caption{Down-up sampling network of the first stage residual attention module.}
 		\label{fig:residualattention_downupsample1}
 	\end{subfigure}
 	\begin{subfigure}{8cm}
 		\centering
 		\includegraphics[width=0.7\linewidth]{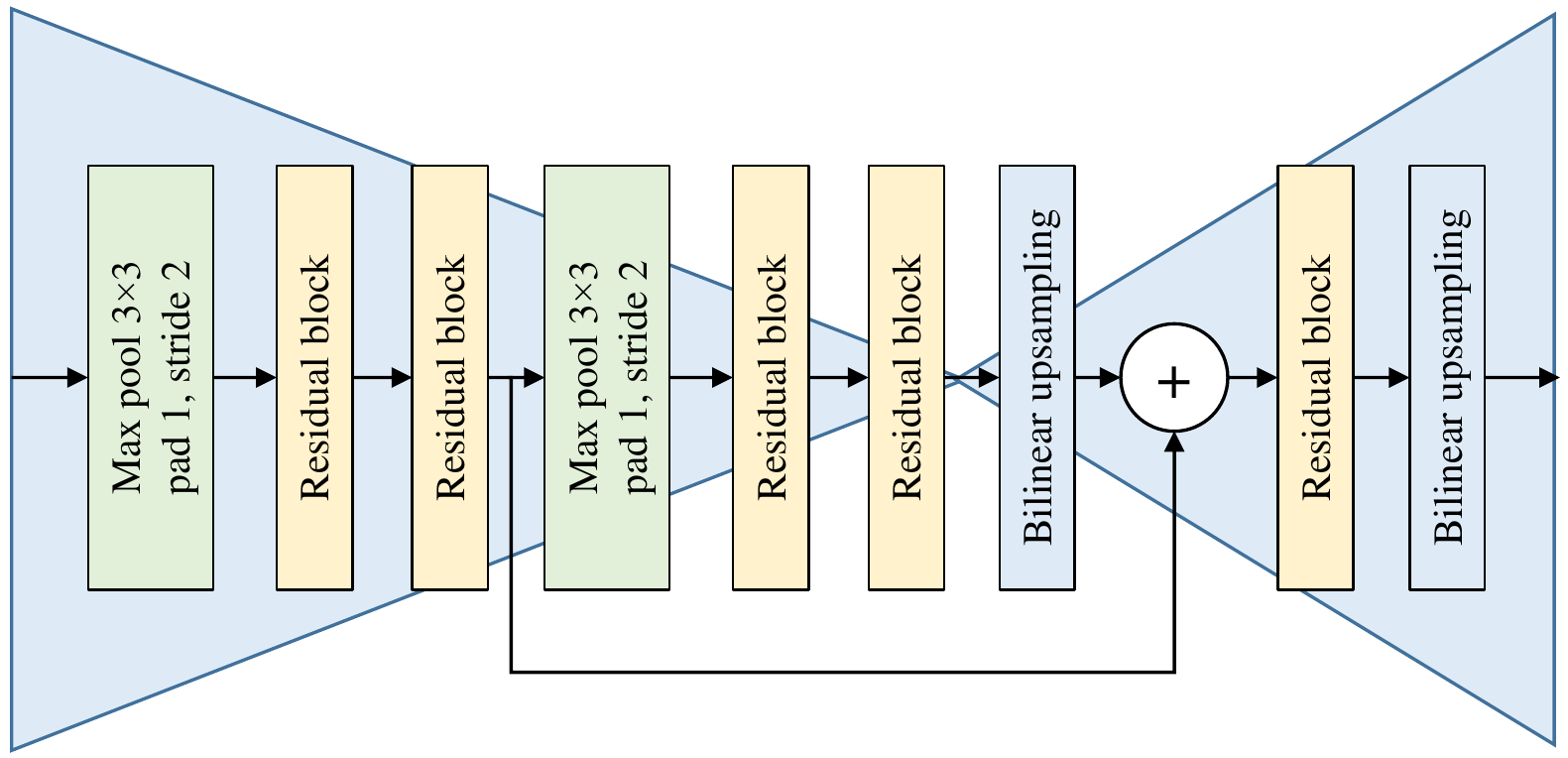}
 		\caption{Down-up sampling network of the second stage residual attention module.}
 		\label{fig:residualattention_downupsample2}
 	\end{subfigure}
 	\begin{subfigure}{8cm}
 		\centering
 		\includegraphics[width=0.5\linewidth]{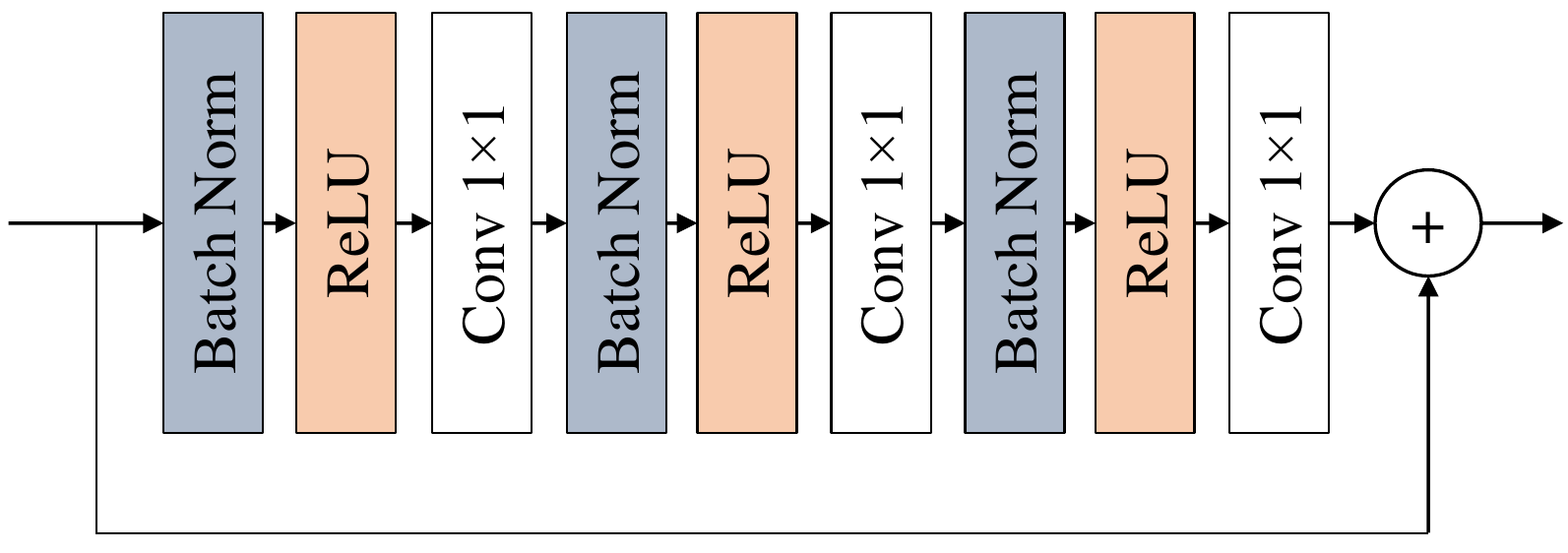}
 		\caption{Residual block.}
 		\label{fig:residual_block}
 	\end{subfigure}
 	\caption{Residual attention module \cite{wang2017residual} and its components}
 	\label{fig:attentionmodule}
 \end{figure}

\subsection{F-SSD: SSD with context by feature fusion} \label{F-SSD}
In order to provide context for a given feature map (target feature) where we want to detect objects, we fuse it with feature maps (context features) from higher layers that the layer of the target features. 
For example in SSD, given our target feature from \verb|conv4_3|, our context features are coming from two layers, they are \verb|conv7| and \verb|conv8_2|, as seen in Fig. \ref{F-SSD}. Although our feature fusion can be generalized to any target feature and any of its higher features.
However, those feature maps have different spatial size, therefore we propose fusion method as described in Fig. \ref{fig:feature_fusion}.
Before fusing by concatenating the features, we perform deconvolution on the context features so they have same spatial size with the target feature. 
We set the context features channels to the half of the target features so the amount of context information is not overwhelming the target features itself.
Just for the F-SSD, we also add one extra convolution layer to the target features that does not change the spatial size and number of channels. 
Furthermore, before concatenating features, a normalization step is very important because each feature values in different layers have different scale. Therefore, we perform batch normalization and ReLU after each layer. 
Finally, we concatenate target features and context features by stacking the features.

\subsection{A-SSD: SSD with attention module} \label{A-SSD}
Visual attention mechanism allows for focusing on part of an image rather than seeing the entire area.
Inspired by the success of residual attention module proposed by Wang et al  \cite{wang2017residual}, we adopt the residual attention module for object detection.
For our A-SSD (Fig. \ref{A-SSD}), we put two-stages residual attention modules after \verb|conv4_3| and  \verb|conv7|. Although it can be generalized to any of layers.
Each of the residual attention stage can be described on Fig. \ref{fig:attentionmodule}. 
It consists of a trunk branch and a mask branch.
The trunk branch has two residual blocks, of each has 3 convolution layers as in Fig. \ref{fig:residual_block}.
The mask branch outputs the attention maps by performing down-sampling and up-sampling with residual connection (Fig. \ref{fig:residualattention_downupsample1} for the first stage and Fig.\ref{fig:residualattention_downupsample2} for the second stage), then finalized with sigmoid activation. 
Residual connections makes the features in down-sampling phase are maintained.
The attention maps from the mask branch are then multiplied with the output of trunk branch, producing attended features.
Finally, the attended features are followed by another residual block, L2 normalization, and ReLU.

\subsection{FA-SSD: Combining feature fusion and attention in SSD} \label{FA-SSD}
We propose method for concatenating two features proposed in section \ref{F-SSD} and \ref{A-SSD}, it can consider context information from the target layer and different layer.
Compare with F-SSD, instead of performing one convolution layer on the target feature, we put one stage attention module, as seen in Fig. \ref{fig:FA-SSD}. The feature fusion method (Fig.\ref{fig:feature_fusion}) is same.


\section{Experiments}

\subsection{Experimental setup}
We applied the proposed method to SSD \cite{liu2016ssd} with same augmentation \footnote{We use models from https://github.com/amdegroot/ssd.pytorch and weights from SSD300 trained on VOC0712 (newest PyTorch weights) https://s3.amazonaws.com/amdegroot-models/ssd300\_mAP\_77.43\_v2.pth for our baseline SSD model. }.
We use SSD with VGG16 backbone and 300 $\times$ 300 input, unless specified otherwise.
For FA-SSD, we applied feature fusion method to \verb|conv4_3| and \verb|conv7| of SSD.
With \verb|conv4_3| as a target, \verb|conv7| and \verb|conv8_2| are used as context layers, and with \verb|conv7| as a target, \verb|conv8_2| and \verb|conv9_2| are used as context layers.
We apply attention module on lower 2 layers for detecting small object. The output of attention module has equal size with target features. 
We trained our models with PASCAL VOC2007 and VOC2012 trainval datasets with learning rate ${10}^{-3}$ for first 80k iterations, then decreased to ${10}^{-4}$ and ${10}^{-5}$ for 100k and 120k iterations, batch size was 16. 
All of test results are tested with VOC2007 test dataset and we follows COCO \cite{lin2014microsoft} for objects size classification, which small objects area is less than 32*32 and large objects area is greater than 96*96. 
We train and test using PyTorch and Titan Xp machine.

\subsection{Ablation studies}
To test on the importance of each feature fusion and attention components compare with SSD baseline, we compare the performance between SSD, F-SSD, A-SSD, and FA-SSD. Table \ref{table_ablation_studies} shows that all F-SSD, A-SSD are better than the SSD which means each components improves the baseline. Although combining fusion and attention as FA-SSD does not show better overall performance compare with F-SSD, FA-SSD shows the best performance and significant improvement on the small objects detection. 

\begin{table}[]
	\centering
	\caption{VOC2007 test results between SSD, F-SSD, A-SSD, and FA-SSD.}
	\label{table_ablation_studies}
	\begin{tabular}{|l|r|r|r|r|r|}
		\hline
		\multirow{2}{*}{} & \multicolumn{1}{c|}{\multirow{2}{*}{mAP}} & \multicolumn{3}{c|}{mAP}                                                              & \multicolumn{1}{l|}{\multirow{2}{*}{FPS}} \\ \cline{3-5}
		& \multicolumn{1}{c|}{}                     & \multicolumn{1}{l|}{Small} & \multicolumn{1}{l|}{Medium} & \multicolumn{1}{l|}{Large} & \multicolumn{1}{l|}{}                     \\ \hline
		SSD  \cite{liu2016ssd}              & 77.5                                      & 20.7                       & 62.0                        & 83.3                       & 23.91                                     \\ \hline
		F-SSD             & \textbf{78.8}                                      & 27.9                       & \textbf{62.8}                       & \textbf{84.1}                       & \textbf{38.14}                                    \\ \hline
		A-SSD             & 78.0                                      & 25.4                       & 62.5                        & 83.5                       & 21.26                                     \\ \hline
		FA-SSD            & 78.1                                      & \textbf{28.5}                       & 61.0                        & 83.6                       & 30.00                                     \\ \hline
	\end{tabular}
\end{table}

\subsection{Inference time}
One interesting thing from results on Table \ref{table_ablation_studies} is that the speed does not always be slower with more components. This motivates us to see the inference time in more detail. Inference time in detection is divided by two, the network inference and the post processing which includes Non-Maximum Suppression (NMS). Based on Table \ref{table_inference_time}, although SSD has the fastest forwarding time, it is the slowest during post processing, hence in total it is still slower than F-SSD and A-SSD.

\begin{table}[]
	\centering
	\caption{Inference time comparison between architectures.}
	\label{table_inference_time}
	\begin{tabular}{|l|r|r|r|}
		\hline
		\multicolumn{1}{|c|}{} & \multicolumn{1}{c|}{\begin{tabular}[c]{@{}c@{}}Total \\time (ms)\end{tabular}}& \multicolumn{1}{c|}{\begin{tabular}[c]{@{}c@{}}Forward \\time (ms)\end{tabular}} & \multicolumn{1}{c|}{\begin{tabular}[c]{@{}c@{}}Post processing \\ time (ms)\end{tabular}} \\ \hline
		SSD  \cite{liu2016ssd}                    & 41.8                          & \textbf{3.8}                                                                              & 37.9                                                                                      \\ \hline
		F-SSD                  & \textbf{26.2}                         & 5.6                                                                              & 20.4                                                                                      \\ \hline
		A-SSD                  & 47.0                          & 22.1                                                                             & 24.9                                                                                      \\ \hline
		FA-SSD                 & 33.3                          & 15.9                                                                             & \textbf{17.3}                                                                                      \\ \hline
	\end{tabular}
\end{table}

\begin{figure}
	\begin{subfigure}{8cm}
		\centering
		\includegraphics[width=\linewidth]{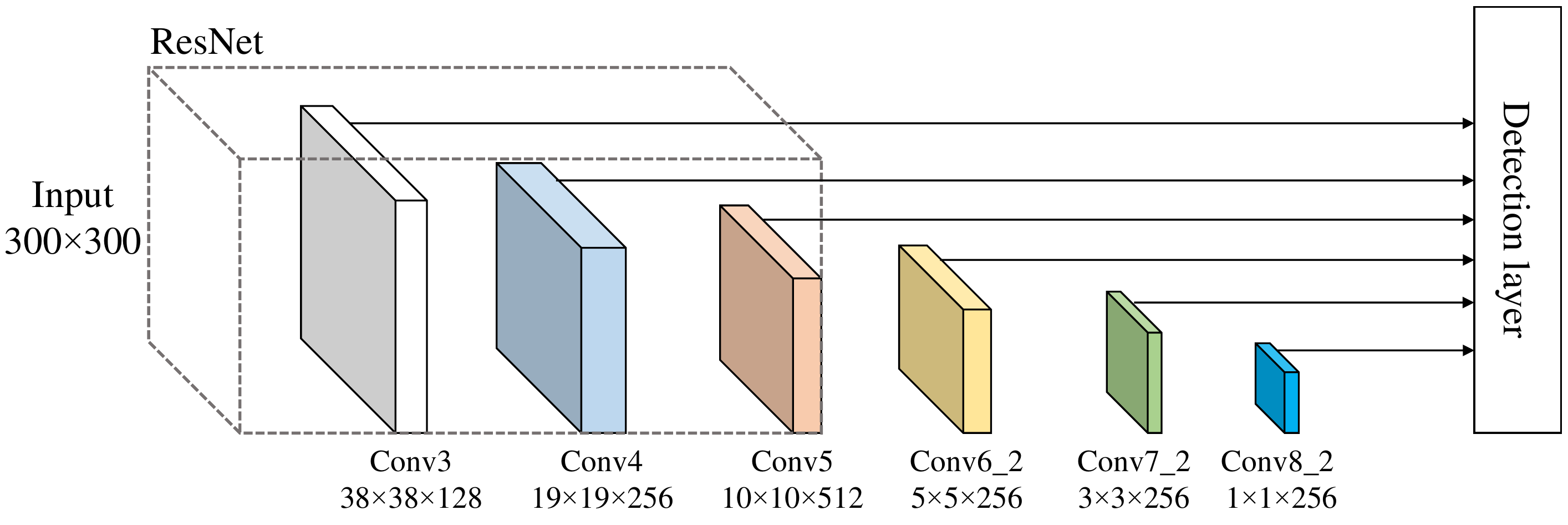}
		\caption{ResNet SSD with input 300$\times$300}
		\label{fig:ResnetSSD}
	\end{subfigure}
	
	\begin{subfigure}{8cm}
		\centering
		\includegraphics[width=\linewidth]{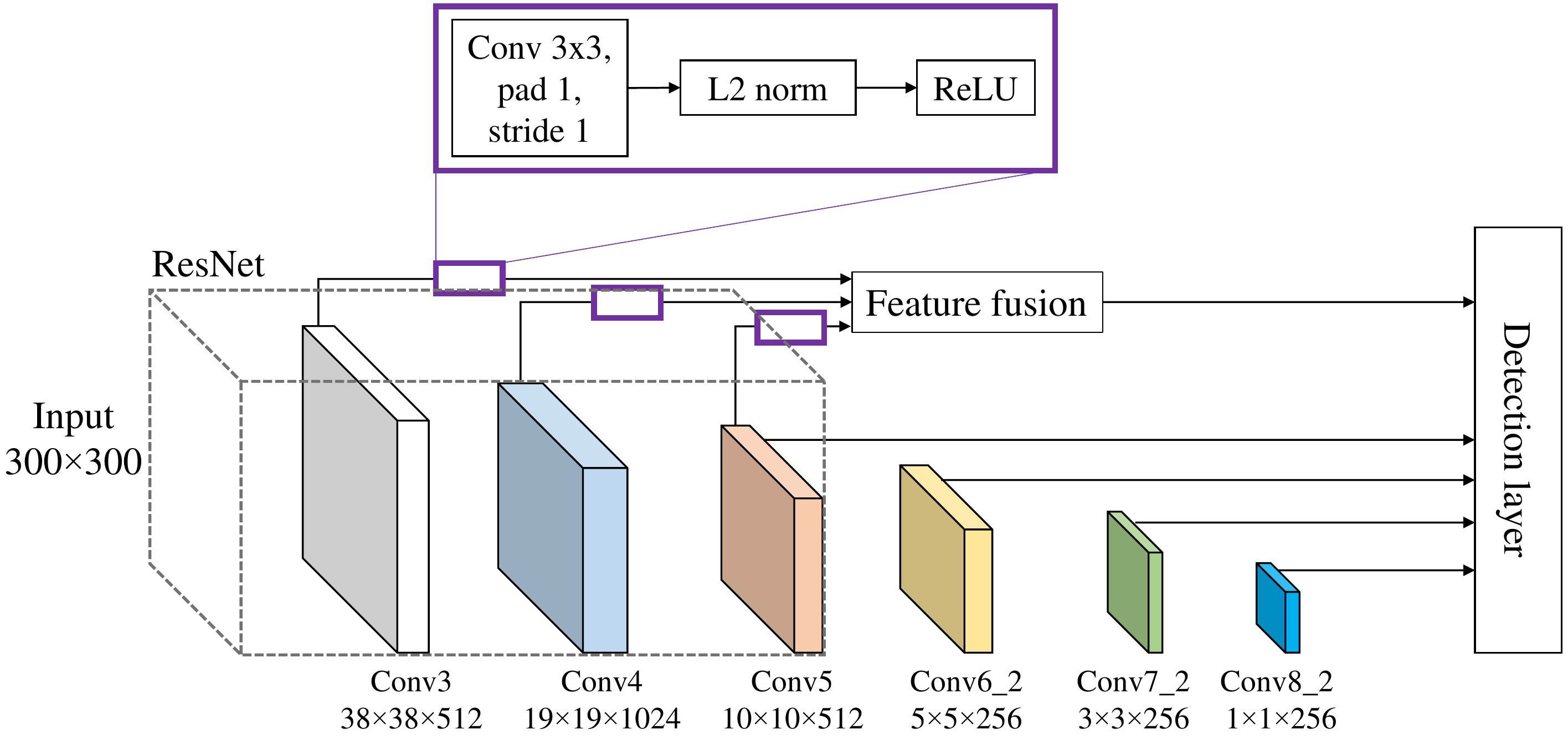}
		\caption{ResNet SSD with feature fusion (F-SSD)}
		\label{fig:ResnetF-SSD}
	\end{subfigure}
	
	\begin{subfigure}{8cm}
		\centering
		\includegraphics[width=\linewidth]{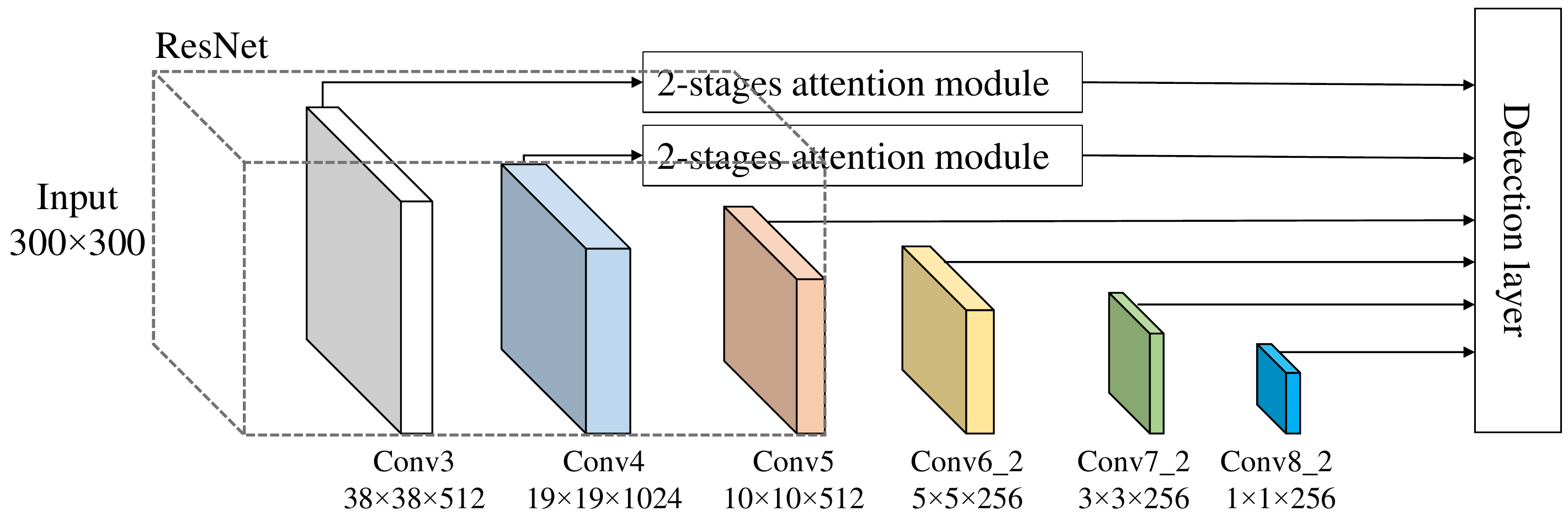}
		\caption{ResNet SSD with attention module (A-SSD)}
		\label{fig:ResnetA-SSD}
	\end{subfigure}
	
	\begin{subfigure}{8cm}
		\centering
		\includegraphics[width=\linewidth]{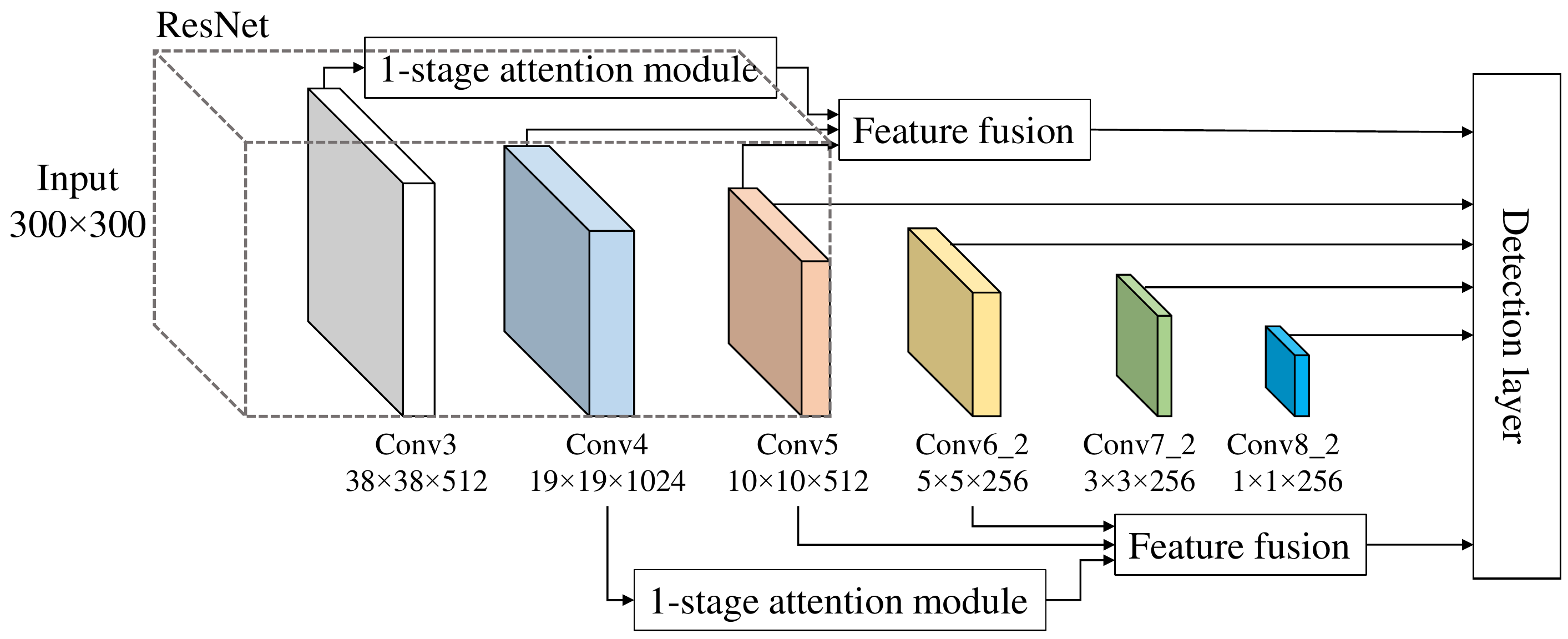}
		\caption{ResNet SSD with feature fusion + attention module (FA-SSD)}
		\label{fig:ResnetFA-SSD}
	\end{subfigure}
	
	\caption{Architectures with ResNet backbone.}\label{fig:architectures_resnet}
\end{figure}

\subsection{Qualitative results}
\begin{figure*}
	\begin{center}
		\includegraphics[width=1\linewidth]{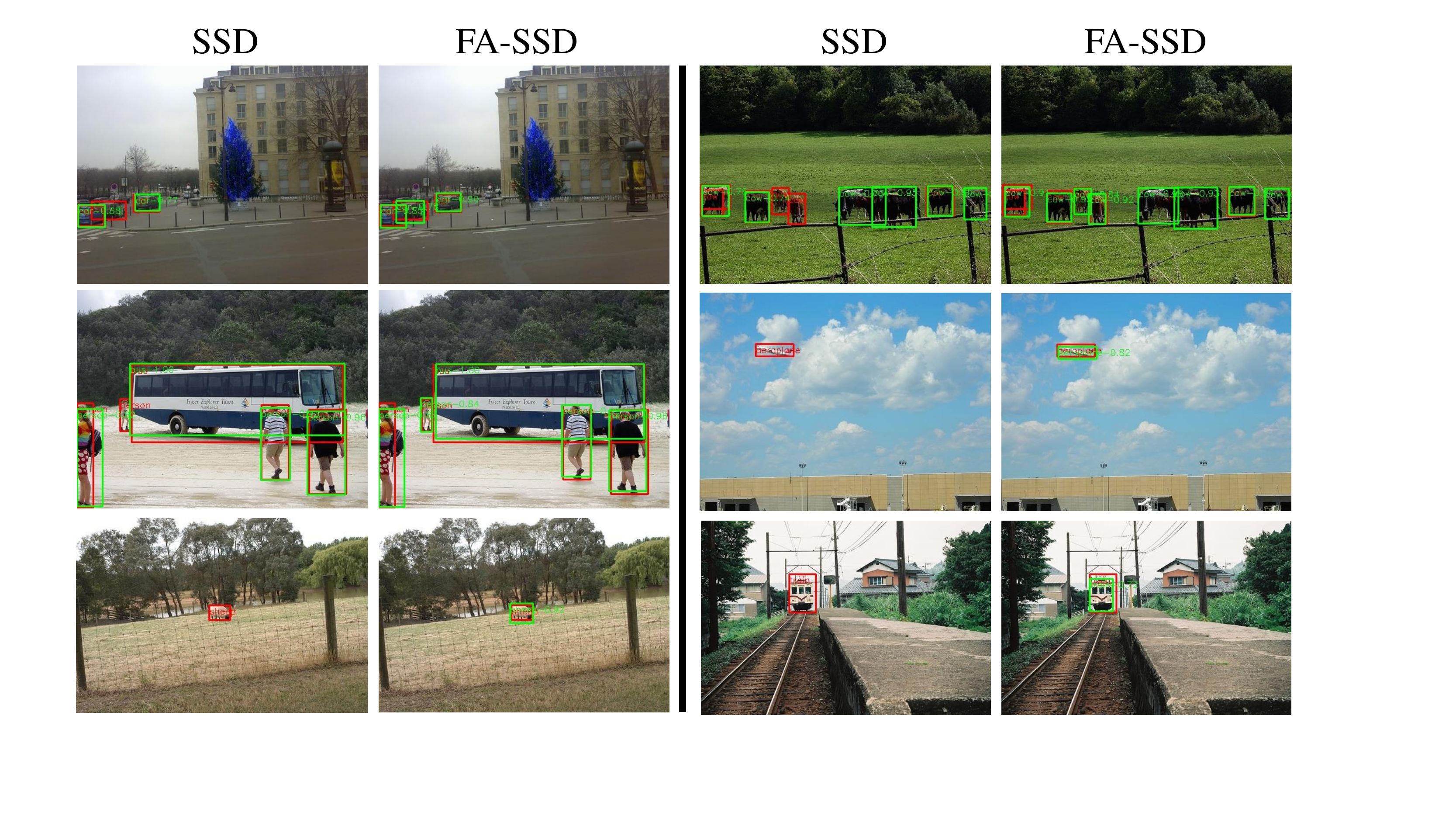}
		\caption{Qualitative results comparison between SSD and FA-SSD. Red box is the ground truth, green box is the prediction.}
		\label{fig:8}
	\end{center}
\end{figure*}

Figure \ref{fig:8} shows the comparison between SSD and FA-SSD qualitatively where SSD fails on detecting small objects when FA-SSD succeeds.

\subsection{Attention visualization}
\begin{figure*}
	\begin{center}
		\includegraphics[width=1\linewidth]{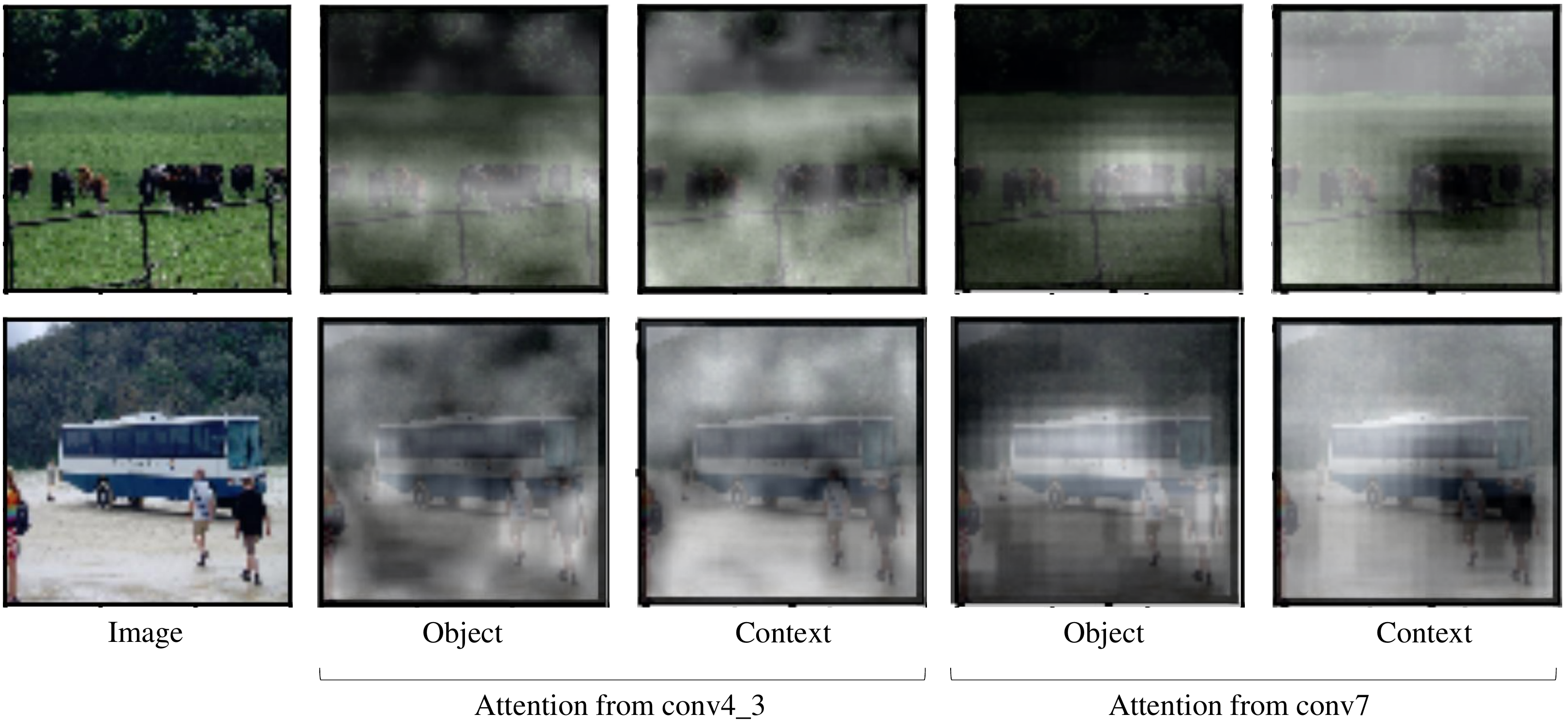}
		\cprotect\caption{Visualization of attention module. Some channels focus on the object and some focus on the context. Attention module on \verb|conv4_3| has higher resolution, therefore can focus on smaller detail compare to attention on  \verb|conv7|.}
		\label{fig:image_attention}
	\end{center}
\end{figure*}

In order to have more understanding on the attention module, we visualize the attention mask from FA-SSD. The attention mask is taken after sigmoid function on Fig. \ref{fig:residualattention}. There are many channels on the attention mask, 512 channels from \verb|conv4_3| and 1024 channels from \verb|conv7|. Each of the channels focuses on different things, both the object and the context. We visualize some samples of the attention masks on Fig. \ref{fig:image_attention}.

\subsection{Generalization on ResNet backbones}
In order to know the generalization with different backbones of SSD, we experiment with ResNet \cite{he2016deep} architectures, specifically ResNet18, ResNet34, and ResNet50. To make the features size same with the original SSD with VGG16 backbone, we take the features from layer 2 results (Fig. \ref{fig:ResnetSSD}). Then F-SSD (Fig. \ref{fig:ResnetF-SSD}), A-SSD (Fig. \ref{fig:ResnetA-SSD}), and FA-SSD (Fig. \ref{fig:ResnetFA-SSD}) just follow the VGG16 backbone version. As seen in Table \ref{table_resnet}, everything follow the trend of the VGG16 backbone version in Table \ref{table_ablation_studies}, except the ResNet34 backbone version does not have the best performance on the small object.

\begin{table}[]
	\centering
	\caption{Results with ResNet backbone architectures. S: small. M: medium. L: large.}
	\label{table_resnet}
	\begin{tabular}{|l|l|r|r|r|r|r|}
		\hline
		\multirow{2}{*}{} & \multirow{2}{*}{Backbone} & \multicolumn{1}{l|}{\multirow{2}{*}{mAP}} & \multicolumn{3}{c|}{mAP}                                                 & \multicolumn{1}{l|}{\multirow{2}{*}{FPS}} \\ \cline{4-6}
		&                          & \multicolumn{1}{l|}{}                     & \multicolumn{1}{c|}{S} & \multicolumn{1}{c|}{M} & \multicolumn{1}{c|}{L} & \multicolumn{1}{l|}{}                     \\ \hline
		SSD               & ResNet18                 & 69.3                                      & 12.6                   & 41.9                   & 80.5                   & \textbf{68.5}                                      \\
		F-SSD             & ResNet18                 & \textbf{74.4}                                      & 12.9                   & \textbf{53.0}                   & \textbf{82.2}                   & 49.5                                      \\
		A-SSD             & ResNet18                 & 73.3                                      & 15.7                   & 51.2                   & 81.6                   & 25.6                                      \\
		FA-SSD            & ResNet18                 & 74.1                                      & \textbf{17.1}                   & 52.8                   & 82.1                   & 33.1                                      \\ \hline
		SSD               & ResNet34                 & 73.4                                      & 11.9                   & 51.3                   & 83.2                   & \textbf{65.7}                                      \\
		F-SSD             & ResNet34                 & \textbf{76.6}                                      & \textbf{16.6}                   & 57.5                   & \textbf{84.4}                   & 53.7                                      \\
		A-SSD             & ResNet34                 & 75.1                                      & 13.0                   & 55.5                   & 83.5                   & 26.7                                      \\
		FA-SSD            & ResNet34                 & 76.5                                      & 14.4                   & \textbf{58.2}                   & 84.2                   & 34.3                                      \\ \hline
		SSD               & ResNet50                 & 74.6                                      & 13.8                   & 52.7                   & 83.7                   & \textbf{54.1}                                      \\
		F-SSD             & ResNet50                 & 77.9                                      & 19.5                   & 60.7                   & 84.5                   & 46.7                                      \\
		A-SSD             & ResNet50                 & 77.8                                      & 16.2                   & 61.3                   & 84.0                   & 26.0                                      \\
		FA-SSD            & ResNet50                 & \textbf{78.3}                                      & \textbf{23.3}                   & \textbf{62.4}                   & \textbf{84.7}                   & 34.7                                      \\ \hline
	\end{tabular}
\end{table}

\subsection{Results on VOC2007 test}
For comparison with other works we compare in Table \ref{table2}. All of the methods compared are trained with VOC2007 trainval and VOC2012 trainval datasets. Although we have lower performance compare to DSSD \cite{fu2017dssd}, our approach runs on 30 FPS while DSSD runs on 12 FPS.

\begin{table}[]
	\small
	\centering
	\caption{Results on PASCAL VOC2007 test}
	\label{table2}
	\begin{tabular}{|l|l|l|}
		\hline
		& Input & mAP           \\ \hline
		YOLO \cite{redmon2016you}          & 448   & 63.4          \\ \hline
		YOLOv2 \cite{redmon2017yolo9000}         & 416   & 76.8          \\ \hline
		Faster R-CNN  \cite{ren2015faster}  &       & 73.2          \\ \hline
		SSD  \cite{liu2016ssd}         & 300   & 77.5          \\ \hline
		DSSD \cite{fu2017dssd}         & 321   & \textbf{78.6} \\ \hline
		FA-SSD (ours) & 300   & {\underline 78.1}    \\ \hline
	\end{tabular}
\end{table}

\section{Conclusion}
In this paper, to improve accuracy for detecting small object, we presented the method for adding context-aware information to Single Shot Multibox Detector. 
Using this method, we can capture context information shown on different layer by fusing multi-scale features and shown on target layer by applying attention mechanism.
Our experiments show improvement in object detection accuracy compared to conventional SSD, especially achieve significantly enhancement for small object.

{\small
\bibliographystyle{ieee}
\bibliography{egbib_1}

\begin{thebibliography}{10}\itemsep=-1pt

\bibitem{everingham2010pascal}
M.~Everingham, L.~Van~Gool, C.~K. Williams, J.~Winn, and A.~Zisserman.
\newblock The pascal visual object classes (voc) challenge.
\newblock {\em International journal of computer vision}, 88(2):303--338, 2010.

\bibitem{fu2017dssd}
C.-Y. Fu, W.~Liu, A.~Ranga, A.~Tyagi, and A.~C. Berg.
\newblock Dssd: Deconvolutional single shot detector.
\newblock {\em arXiv preprint arXiv:1701.06659}, 2017.

\bibitem{girshick2015fast}
R.~Girshick.
\newblock Fast r-cnn.
\newblock {\em arXiv preprint arXiv:1504.08083}, 2015.

\bibitem{girshick2014rich}
R.~Girshick, J.~Donahue, T.~Darrell, and J.~Malik.
\newblock Rich feature hierarchies for accurate object detection and semantic
  segmentation.
\newblock In {\em Proceedings of the IEEE conference on computer vision and
  pattern recognition}, pages 580--587, 2014.

\bibitem{goodfellow2014generative}
I.~Goodfellow, J.~Pouget-Abadie, M.~Mirza, B.~Xu, D.~Warde-Farley, S.~Ozair,
  A.~Courville, and Y.~Bengio.
\newblock Generative adversarial nets.
\newblock In {\em Advances in neural information processing systems}, pages
  2672--2680, 2014.

\bibitem{he2016deep}
K.~He, X.~Zhang, S.~Ren, and J.~Sun.
\newblock Deep residual learning for image recognition.
\newblock In {\em Proceedings of the IEEE conference on computer vision and
  pattern recognition}, pages 770--778, 2016.

\bibitem{jeong2017enhancement}
J.~Jeong, H.~Park, and N.~Kwak.
\newblock Enhancement of ssd by concatenating feature maps for object
  detection.
\newblock {\em arXiv preprint arXiv:1705.09587}, 2017.

\bibitem{li2017perceptual}
J.~Li, X.~Liang, Y.~Wei, T.~Xu, J.~Feng, and S.~Yan.
\newblock Perceptual generative adversarial networks for small object
  detection.
\newblock In {\em IEEE CVPR}, 2017.

\bibitem{lin2014microsoft}
T.-Y. Lin, M.~Maire, S.~Belongie, J.~Hays, P.~Perona, D.~Ramanan,
  P.~Doll{\'a}r, and C.~L. Zitnick.
\newblock Microsoft coco: Common objects in context.
\newblock In {\em European conference on computer vision}, pages 740--755.
  Springer, 2014.

\bibitem{liu2016ssd}
W.~Liu, D.~Anguelov, D.~Erhan, C.~Szegedy, S.~Reed, C.-Y. Fu, and A.~C. Berg.
\newblock Ssd: Single shot multibox detector.
\newblock In {\em European conference on computer vision}, pages 21--37.
  Springer, 2016.

\bibitem{redmon2016you}
J.~Redmon, S.~Divvala, R.~Girshick, and A.~Farhadi.
\newblock You only look once: Unified, real-time object detection.
\newblock In {\em Proceedings of the IEEE conference on computer vision and
  pattern recognition}, pages 779--788, 2016.

\bibitem{redmon2017yolo9000}
J.~Redmon and A.~Farhadi.
\newblock Yolo9000: better, faster, stronger.
\newblock {\em arXiv preprint}, 2017.

\bibitem{ren2015faster}
S.~Ren, K.~He, R.~Girshick, and J.~Sun.
\newblock Faster r-cnn: Towards real-time object detection with region proposal
  networks.
\newblock In {\em Advances in neural information processing systems}, pages
  91--99, 2015.

\bibitem{sharma2015action}
S.~Sharma, R.~Kiros, and R.~Salakhutdinov.
\newblock Action recognition using visual attention.
\newblock {\em arXiv preprint arXiv:1511.04119}, 2015.

\bibitem{simonyan2014very}
K.~Simonyan and A.~Zisserman.
\newblock Very deep convolutional networks for large-scale image recognition.
\newblock {\em arXiv preprint arXiv:1409.1556}, 2014.

\bibitem{uijlings2013selective}
J.~R. Uijlings, K.~E. Van De~Sande, T.~Gevers, and A.~W. Smeulders.
\newblock Selective search for object recognition.
\newblock {\em International journal of computer vision}, 104(2):154--171,
  2013.

\bibitem{wang2017residual}
F.~Wang, M.~Jiang, C.~Qian, S.~Yang, C.~Li, H.~Zhang, X.~Wang, and X.~Tang.
\newblock Residual attention network for image classification.
\newblock {\em arXiv preprint arXiv:1704.06904}, 2017.

\bibitem{xu2015show}
K.~Xu, J.~Ba, R.~Kiros, K.~Cho, A.~Courville, R.~Salakhudinov, R.~Zemel, and
  Y.~Bengio.
\newblock Show, attend and tell: Neural image caption generation with visual
  attention.
\newblock In {\em International conference on machine learning}, pages
  2048--2057, 2015.

\end{thebibliography}
}

\newpage
\appendix
\section{Detail inference time on ResNet backbones}
Table \ref{table_resnet_time} shows the detail on inference time for the ResNet backbone architectures.
\begin{table}[]
	\centering
	\caption{Time detail on ResNet backbone versions.}
	\label{table_resnet_time}
	\begin{tabular}{|l|l|r|r|r|}
		\hline
		\multicolumn{1}{|c|}{} & \multicolumn{1}{c|}{Network} & \multicolumn{1}{c|}{\begin{tabular}[c]{@{}c@{}}Total\\ time (ms)\end{tabular}} & \multicolumn{1}{c|}{\begin{tabular}[c]{@{}c@{}}Forward\\ time (ms)\end{tabular}} & \multicolumn{1}{c|}{\begin{tabular}[c]{@{}c@{}}Post processing\\ time (ms)\end{tabular}} \\ \hline
		SSD                    & Resnet18                     & 14.6                                                                           & 4.6                                                                                      & 10.3                                                                                     \\ 
		F-SSD                  & Resnet18                     & 20.2                                                                           & 6.1                                                                                      & 14.1                                                                                     \\ 
		A-SSD                  & Resnet18                     & 39.1                                                                           & 23.0                                                                                     & 16.0                                                                                     \\ 
		FA-SSD                 & Resnet18                     & 30.2                                                                           & 16.5                                                                                     & 13.7                                                                                     \\ \hline
		SSD                    & Resnet34                     & 15.2                                                                           & 6.1                                                                                      & 9.0                                                                                      \\ 
		F-SSD                  & Resnet34                     & 18.6                                                                           & 8.0                                                                                      & 10.8                                                                                     \\
		A-SSD                  & Resnet34                     & 37.4                                                                           & 24.1                                                                                     & 13.3                                                                                     \\
		FA-SSD                 & Resnet34                     & 29.2                                                                           & 18.1                                                                                     & 11.0                                                                                     \\ \hline
		SSD                    & Resnet50                     & 18.5                                                                           & 8.8                                                                                      & 9.6                                                                                      \\
		F-SSD                  & Resnet50                     & 21.4                                                                           & 9.7                                                                                      & 11.6                                                                                     \\
		A-SSD                  & Resnet50                     & 38.4                                                                           & 27.4                                                                                     & 11.0                                                                                     \\
		FA-SSD                 & Resnet50                     & 28.8                                                                           & 19.1                                                                                     & 9.7                                                                                      \\ \hline
	\end{tabular}
\end{table}

\section{VOC2012 test results}
Table \ref{table_voc2012} shows the FA-SSD does not improve the SSD. The reason needs to be investigated further such as the distribution of object size of VOC2012. Especially, FA-SSD based on Table \ref{table_ablation_studies} actually has degradation on medium size object compare to SSD.
\begin{table*}[]
    \centering
	\scriptsize
	\caption{Results on VOC2012 test.}
	\label{table_voc2012}
	\renewcommand\tabcolsep{2pt}
	\begin{tabular}{|l|r|r|r|r|r|r|r|r|r|r|r|r|r|r|r|r|r|r|r|r|r|}
		\hline
		Method       & \multicolumn{1}{c|}{mAP} & \multicolumn{1}{c|}{aero} & \multicolumn{1}{c|}{bike} & \multicolumn{1}{c|}{bird} & \multicolumn{1}{c|}{boat} & \multicolumn{1}{c|}{bottle} & \multicolumn{1}{c|}{bus} & \multicolumn{1}{c|}{car} & \multicolumn{1}{c|}{cat} & \multicolumn{1}{c|}{chair} & \multicolumn{1}{c|}{cow} & \multicolumn{1}{c|}{table} & \multicolumn{1}{c|}{dog} & \multicolumn{1}{c|}{horse} & \multicolumn{1}{c|}{mbike} & \multicolumn{1}{c|}{person} & \multicolumn{1}{c|}{plant} & \multicolumn{1}{c|}{sheep} & \multicolumn{1}{c|}{sofa} & \multicolumn{1}{c|}{train} & \multicolumn{1}{c|}{tv} \\ \hline
		Faster R-CNN & 73.2                     & 76.5                      & 79.0                      & 70.9                      & 65.5                      & \textbf{52.1}               & \textbf{83.1}            & \textbf{84.7}            & 86.4                     & 52.0                       & \textbf{81.9}            & 65.7                       & 84.8                     & 84.6                       & 77.5                       & 76.7                        & 38.8                       & 73.6                       & 73.9                      & 83.0                       & 72.6                    \\ \hline
		YOLO         & 57.9                     & 77                        & 67.2                      & 57.7                      & 38.3                      & 22.7                        & 68.3                     & 55.9                     & 81.4                     & 36.2                       & 60.8                     & 48.5                       & 77.2                     & 72.3                       & 71.3                       & 63.5                        & 28.9                       & 52.2                       & 54.8                      & 73.9                       & 50.8                    \\ \hline
		YOLOv2 544   & 73.4                     & 86.3                      & 82.0                      & \textbf{74.8}             & 59.2                      & 51.8                        & 79.8                     & 76.5                     & \textbf{90.6}            & 52.1                       & 78.2                     & 58.5                       & \textbf{89.3}            & 82.5                       & 83.4                       & \textbf{81.3}               & \textbf{49.1}              & 77.2                       & 62.4                      & 83.8                       & 68.7                    \\ \hline
		SSD          & \textbf{74.3}            & 75.5                      & 80.2                      & 72.3                      & \textbf{66.3}             & 47.6                        & 83.0                     & 84.2                     & 86.1                     & 54.7                       & 78.3                     & \textbf{73.9}              & 84.5                     & \textbf{85.3}              & 82.6                       & 76.2                        & 48.6                       & 73.9                       & \textbf{76.0}             & 83.4                       & \textbf{74.0}           \\ \hline
		FA-SSD       & 74.1                     & \textbf{87.2}             & \textbf{82.7}             & 73.1                      & 62.0                      & 49.4                        & 82.1                     & 76.2                     & 89.0                     & \textbf{55.4}              & 77.7                     & 61.8                       & 87.1                     & 85.0                       & \textbf{84.2}              & 80.7                        & 48.9                       & \textbf{78.5}              & 67.0                      & \textbf{84.7}              & 69.2                    \\ \hline
	\end{tabular}
\end{table*}

\section{Detail classes results on VOC2007}
Table \ref{table_detail_class} shows the mAP from VOC2007 test data for each classes of every architectures.
\begin{table*}[]
	\centering
	\scriptsize
	\caption{Detail mAP for every classes in every architectures on VOC2007. No result means no object with the respective size.}
	\label{table_detail_class}
	\renewcommand\tabcolsep{2pt}
	\begin{tabular}{|l|l|r|r|r|r|r|r|r|r|r|r|r|r|r|r|r|r|r|r|r|r|}
		\hline
		\multirow{2}{*}{}                       & \multicolumn{1}{c|}{\multirow{2}{*}{Network}} & \multicolumn{4}{c|}{Aeroplane}                                                                      & \multicolumn{4}{c|}{Bicycle}                                                                        & \multicolumn{4}{c|}{Bird}                                                                           & \multicolumn{4}{c|}{Boat}                                                                           & \multicolumn{4}{c|}{Bottle}                                                                         \\ \cline{3-22} 
		& \multicolumn{1}{c|}{}                         & \multicolumn{1}{c|}{mAP} & \multicolumn{1}{c|}{S} & \multicolumn{1}{c|}{M} & \multicolumn{1}{c|}{L} & \multicolumn{1}{c|}{mAP} & \multicolumn{1}{c|}{S} & \multicolumn{1}{c|}{M} & \multicolumn{1}{c|}{L} & \multicolumn{1}{c|}{mAP} & \multicolumn{1}{c|}{S} & \multicolumn{1}{c|}{M} & \multicolumn{1}{c|}{L} & \multicolumn{1}{c|}{mAP} & \multicolumn{1}{c|}{S} & \multicolumn{1}{c|}{M} & \multicolumn{1}{c|}{L} & \multicolumn{1}{c|}{mAP} & \multicolumn{1}{c|}{S} & \multicolumn{1}{c|}{M} & \multicolumn{1}{c|}{L} \\ \hline
		SSD                                     & VGG16                                         & 82.1                     & 32.4                   & 80.0                   & \textbf{89.4}          & 85.7                     & 14.3                   & 75.8                   & 88.5                   & 75.5                     & 19.2                   & 65.1                   & 83.1                   & 69.5                     & \textbf{27.2}          & 59.0                   & 80.1                   & 50.2                     & \textbf{10.5}          & 48.3                   & \textbf{67.8}          \\
		F-SSD                                   & VGG16                                         & 82.0                     & 35.5                   & 78.9                   & 88.3                   & \textbf{87.1}            & \textbf{16.7}          & 76.0                   & \textbf{89.4}          & \textbf{76.8}            & 14.3                   & \textbf{68.5}          & 85.2                   & \textbf{73.6}            & 22.5                   & \textbf{64.7}          & 80.6                   & \textbf{54.7}            & 6.6                    & 55.3                   & 72.4                   \\
		A-SSD                                   & VGG16                                         & \textbf{82.8}            & 32.1                   & \textbf{82.1}          & 89.0                   & 85.9                     & 14.3                   & \textbf{79.0}          & 88.0                   & 77.6                     & 23.0                   & 67.7                   & 87.4                   & 73.0                     & 15.7                   & 61.8                   & \textbf{82.3}          & 51.5                     & 9.5                    & 52.3                   & 67.1                   \\
		FA-SSD                                  & VGG16                                         & 80.3                     & \textbf{37.2}          & 76.6                   & 86.3                   & 84.9                     & \textbf{16.7}          & 71.8                   & 88.0                   & 76.7                     & \textbf{28.8}          & 62.5                   & \textbf{86.8}          & 70.4                     & 24.4                   & 59.5                   & 80.3                   & 53.5                     & 9.7                    & \textbf{54.3}          & 66.2                   \\ \hline
		SSD                                     & Resnet18                                      & 68.9                     & \textbf{18.2}          & 55.4                   & 86.7                   & 77.4                     & 33.3                   & 59.9                   & 80.5                   & 64.1                     & 15.2                   & 29.6                   & 79.2                   & 58.8                     & 4.2                    & 34.6                   & 75.4                   & 37.9                     & 0.8                    & 30.3                   & 67.4                   \\
		F-SSD                                   & Resnet18                                      & 76.2                     & 7.6                    & \textbf{73.0}          & \textbf{86.8}          & 83.2                     & 14.3                   & \textbf{72.9}          & 85.9                   & \textbf{72.7}            & 15.6                   & \textbf{56.2}          & \textbf{83.7}          & 67.2                     & \textbf{20.1}          & \textbf{53.9}          & 78.7                   & \textbf{43.5}            & 3.5                    & \textbf{40.0}          & 69.0                   \\
		A-SSD                                   & Resnet18                                      & \textbf{76.6}            & 12.7                   & 72.8                   & 86.5                   & 81.6                     & 20.0                   & 66.4                   & 85.9                   & 67.3                     & \textbf{23.4}          & 45.6                   & 78.7                   & 65.7                     & 13.4                   & 45.3                   & \textbf{80.8}          & 39.7                     & \textbf{12.0}          & 33.3                   & 65.6                   \\
		FA-SSD                                  & Resnet18                                      & 75.7                     & 14.3                   & 69.9                   & 85.3                   & \textbf{83.9}            & \textbf{50.0}          & 70.0                   & \textbf{87.0}          & 71.2                     & 20.2                   & 52.4                   & 79.5                   & \textbf{67.4}            & 13.2                   & 51.0                   & 80.2                   & 42.5                     & 1.7                    & 38.4                   & \textbf{69.7}          \\ \hline
		SSD                                     & Resnet34                                      & 76.5                     & \textbf{14.6}          & 67.6                   & 89.0                   & 80.1                     & 0.0                    & 75.7                   & 87.2                   & 68.6                     & 14.6                   & 44.7                   & 79.8                   & 68.1                     & 14.9                   & 50.2                   & 78.8                   & 43.0                     & 3.7                    & 36.3                   & 74.2                   \\
		F-SSD                                   & Resnet34                                      & \textbf{78.3}            & 12.4                   & \textbf{75.1}          & \textbf{89.1}          & \textbf{86.2}            & \textbf{50.0}          & 76.1                   & \textbf{88.7}          & 75.4                     & 22.0                   & 60.7                   & 86.1                   & 67.9                     & 14.9                   & 51.7                   & 80.9                   & 48.0                     & 1.6                    & 44.8                   & 74.0                   \\
		A-SSD                                   & Resnet34                                      & 76.5                     & 13.4                   & 70.7                   & 86.7                   & 84.0                     & 0.0                    & \textbf{76.5}          & 87.9                   & 73.8                     & \textbf{22.2}          & 54.1                   & \textbf{86.7}          & 65.6                     & \textbf{24.5}          & 44.7                   & 78.4                   & 44.9                     & \textbf{10.4}          & 38.4                   & 74.2                   \\
		FA-SSD                                  & Resnet34                                      & 77.5                     & 10.4                   & 72.8                   & 88.3                   & 84.9                     & 0.0                    & 74.7                   & 88.5                   & \textbf{75.8}            & 22.0                   & \textbf{61.8}          & \textbf{86.7}          & \textbf{70.3}            & 12.0                   & \textbf{54.2}          & \textbf{83.4}          & \textbf{50.0}            & 1.1                    & \textbf{45.9}          & \textbf{76.2}          \\ \hline
		SSD                                     & Resnet50                                      & 71.0                     & 18.2                   & 65.9                   & 86.5                   & 85.7                     & 25.0                   & 74.4                   & \textbf{88.9}          & 69.3                     & 21.8                   & 45.7                   & 87.3                   & 64.2                     & 20.7                   & 49.1                   & 77.1                   & 42.2                     & 10.3                   & 37.6                   & 69.8                   \\
		F-SSD                                   & Resnet50                                      & \textbf{79.8}            & \textbf{34.0}          & \textbf{84.2}          & 88.1                   & \textbf{86.8}            & 33.3                   & \textbf{76.9}          & 88.5                   & 78.1                     & 23.3                   & 61.8                   & \textbf{88.6}          & 71.1                     & \textbf{24.3}          & 59.2                   & 78.4                   & 50.8                     & 4.0                    & 44.8                   & \textbf{77.8}          \\
		A-SSD                                   & Resnet50                                      & 79.2                     & 18.8                   & 82.8                   & 88.9                   & 86.3                     & 20.0                   & 76.6                   & 88.4                   & 76.5                     & \textbf{25.5}          & 60.5                   & 87.3                   & 67.7                     & 16.2                   & \textbf{62.3}          & 75.1                   & 47.1                     & \textbf{11.7}          & 44.6                   & 65.5                   \\
		FA-SSD                                  & Resnet50                                      & \textbf{79.8}            & 22.0                   & 79.1                   & \textbf{89.1}          & 80.9                     & \textbf{100.0}         & 74.6                   & 88.6                   & \textbf{78.2}            & 20.8                   & \textbf{66.3}          & 88.0                   & \textbf{71.8}            & 21.3                   & 61.9                   & \textbf{81.3}          & \textbf{51.1}            & 5.4                    & \textbf{49.2}          & 71.8                   \\ \hline
		\multicolumn{1}{|c|}{\multirow{2}{*}{}} & \multicolumn{1}{c|}{\multirow{2}{*}{Network}} & \multicolumn{4}{c|}{Bus}                                                                            & \multicolumn{4}{c|}{Car}                                                                            & \multicolumn{4}{c|}{Cat}                                                                            & \multicolumn{4}{c|}{Chair}                                                                          & \multicolumn{4}{c|}{Cow}                                                                            \\ \cline{3-22} 
		\multicolumn{1}{|c|}{}                  & \multicolumn{1}{c|}{}                         & \multicolumn{1}{c|}{mAP} & \multicolumn{1}{c|}{S} & \multicolumn{1}{c|}{M} & \multicolumn{1}{c|}{L} & \multicolumn{1}{c|}{mAP} & \multicolumn{1}{c|}{S} & \multicolumn{1}{c|}{M} & \multicolumn{1}{c|}{L} & \multicolumn{1}{c|}{mAP} & \multicolumn{1}{c|}{S} & \multicolumn{1}{c|}{M} & \multicolumn{1}{c|}{L} & \multicolumn{1}{c|}{mAP} & \multicolumn{1}{c|}{S} & \multicolumn{1}{c|}{M} & \multicolumn{1}{c|}{L} & \multicolumn{1}{c|}{mAP} & \multicolumn{1}{c|}{S} & \multicolumn{1}{c|}{M} & \multicolumn{1}{c|}{L} \\ \hline
		SSD                                     & VGG16                                         & 84.8                     & 8.3                    & 59.2                   & 88.5                   & 85.8                     & 31.2                   & 81.2                   & \textbf{90.6}          & 87.3                     & 0.0                    & 56.9                   & 89.3                   & 61.4                     & 0.0                    & 54.0                   & 67.2                   & 82.4                     & 17.3                   & 78.5                   & 87.1                   \\
		F-SSD                                   & VGG16                                         & \textbf{87.2}            & \textbf{100.0}         & 53.7                   & 89.6                   & 86.4                     & 33.1                   & 81.0                   & 90.5                   & 89.0                     & 0.0                    & \textbf{60.4}          & 89.6                   & 61.7                     & 0.0                    & 52.3                   & 69.5                   & \textbf{86.3}            & \textbf{44.7}          & 83.6                   & \textbf{88.0}          \\
		A-SSD                                   & VGG16                                         & 86.9                     & \textbf{100.0}         & \textbf{61.0}          & 89.4                   & 86.4                     & 34.9                   & \textbf{82.6}          & 90.5                   & 88.5                     & 0.0                    & 57.0                   & 89.5                   & 62.5                     & 0.0                    & 52.8                   & 69.4                   & 82.8                     & 24.8                   & 77.9                   & 86.0                   \\
		FA-SSD                                  & VGG16                                         & 86.0                     & \textbf{100.0}         & 51.0                   & \textbf{89.9}          & \textbf{86.9}            & \textbf{33.8}          & 82.3                   & \textbf{90.6}          & \textbf{89.8}            & 0.0                    & 56.0                   & \textbf{90.2}          & \textbf{63.2}            & 0.0                    & \textbf{54.6}          & \textbf{70.3}          & 85.2                     & 38.1                   & \textbf{83.7}          & 86.3                   \\ \hline
		SSD                                     & Resnet18                                      & 78.6                     & 0.0                    & \textbf{46.5}          & 88.3                   & 77.6                     & 9.3                    & 69.0                   & 90.3                   & 85.9                     & 0.0                    & 24.8                   & 88.5                   & 52.3                     & 0.0                    & 34.6                   & 64.2                   & 71.0                     & \textbf{45.5}          & 55.7                   & 83.4                   \\
		F-SSD                                   & Resnet18                                      & 81.6                     & 0.0                    & 42.9                   & 88.4                   & 82.5                     & 17.5                   & 75.0                   & \textbf{90.5}          & 85.4                     & 0.0                    & \textbf{46.8}          & 88.6                   & 57.9                     & 0.8                    & 40.5                   & 69.1                   & \textbf{79.6}            & 13.6                   & \textbf{73.0}          & \textbf{85.0}          \\
		A-SSD                                   & Resnet18                                      & 82.8                     & 0.0                    & 41.4                   & \textbf{89.3}          & 81.6                     & 19.5                   & 75.0                   & 90.4                   & 86.2                     & 0.0                    & 39.1                   & 88.5                   & 57.8                     & 0.0                    & 42.5                   & 68.5                   & 76.1                     & 31.3                   & 68.3                   & 82.4                   \\
		FA-SSD                                  & Resnet18                                      & 79.0                     & 0.0                    & 36.3                   & 88.8                   & \textbf{82.7}            & \textbf{19.9}          & \textbf{75.2}          & 90.4                   & \textbf{86.3}            & 0.0                    & 37.6                   & \textbf{89.1}          & \textbf{60.2}            & \textbf{1.5}           & \textbf{45.6}          & \textbf{69.5}          & 78.6                     & 34.1                   & \textbf{73.0}          & 83.0                   \\ \hline
		SSD                                     & Resnet34                                      & 79.2                     & 0.0                    & 42.2                   & 88.9                   & 79.4                     & 17.8                   & 73.6                   & 90.4                   & 80.5                     & 0.0                    & 37.7                   & 88.8                   & 58.7                     & 0.0                    & 41.2                   & 70.8                   & 74.3                     & 22.7                   & 65.8                   & 84.1                   \\
		F-SSD                                   & Resnet34                                      & 79.9                     & 0.0                    & 39.9                   & \textbf{89.2}          & 83.8                     & 22.9                   & 77.6                   & \textbf{90.6}          & 86.0                     & 0.0                    & 50.6                   & 87.7                   & \textbf{63.1}            & 2.0                    & 50.5                   & \textbf{71.2}          & 81.3                     & \textbf{30.5}          & 73.9                   & 86.7                   \\
		A-SSD                                   & Resnet34                                      & \textbf{82.8}            & 0.0                    & \textbf{46.9}          & 88.8                   & 84.3                     & 22.2                   & \textbf{78.3}          & 90.5                   & \textbf{87.1}            & 0.0                    & 48.9                   & \textbf{89.0}          & 59.9                     & 0.0                    & 45.2                   & 69.1                   & 80.6                     & 17.1                   & 73.7                   & 87.3                   \\
		FA-SSD                                  & Resnet34                                      & 79.9                     & 0.0                    & 46.1                   & \textbf{89.2}          & \textbf{85.1}            & \textbf{26.2}          & 75.8                   & 90.5                   & 87.0                     & 0.0                    & \textbf{51.5}          & 88.6                   & 62.3                     & \textbf{3.7}           & \textbf{50.7}          & 70.3                   & \textbf{83.0}            & 11.7                   & \textbf{77.8}          & \textbf{88.2}          \\ \hline
		SSD                                     & Resnet50                                      & \textbf{86.1}            & 0.0                    & 49.0                   & 89.7                   & 79.8                     & 17.2                   & 74.3                   & 90.6                   & 88.3                     & 0.0                    & 42.8                   & \textbf{89.7}          & 59.4                     & 0.0                    & 45.0                   & 69.9                   & 80.1                     & 11.5                   & 67.7                   & \textbf{87.2}          \\
		F-SSD                                   & Resnet50                                      & 80.7                     & \textbf{33.3}          & 49.6                   & \textbf{89.8}          & 85.8                     & 26.0                   & 81.0                   & 90.6                   & \textbf{88.6}            & 0.0                    & 57.5                   & 89.4                   & 63.1                     & 0.0                    & 51.7                   & 70.0                   & \textbf{83.8}            & \textbf{25.7}          & \textbf{80.6}          & \textbf{87.2}          \\
		A-SSD                                   & Resnet50                                      & 86.0                     & 0.0                    & \textbf{60.3}          & 89.7                   & 86.1                     & \textbf{29.2}          & 82.2                   & \textbf{90.7}          & 87.0                     & 0.0                    & 47.9                   & 88.7                   & \textbf{64.6}            & 3.3                    & 52.9                   & \textbf{72.1}          & 81.4                     & 3.6                    & 75.9                   & 85.8                   \\
		FA-SSD                                  & Resnet50                                      & 85.8                     & 20.0                   & 53.8                   & 89.7                   & \textbf{86.6}            & 25.4                   & \textbf{82.3}          & 90.6                   & 88.3                     & 0.0                    & \textbf{61.8}          & 89.0                   & 63.7                     & \textbf{3.7}           & \textbf{53.7}          & 70.1                   & 83.6                     & 21.6                   & 79.7                   & 86.9                   \\ \hline
		\multicolumn{1}{|c|}{\multirow{2}{*}{}} & \multicolumn{1}{c|}{\multirow{2}{*}{Network}} & \multicolumn{4}{c|}{Dining table}                                                                   & \multicolumn{4}{c|}{Dog}                                                                            & \multicolumn{4}{c|}{Horse}                                                                          & \multicolumn{4}{c|}{Motorbike}                                                                      & \multicolumn{4}{c|}{Person}                                                                         \\ \cline{3-22} 
		\multicolumn{1}{|c|}{}                  & \multicolumn{1}{c|}{}                         & \multicolumn{1}{c|}{mAP} & \multicolumn{1}{c|}{S} & \multicolumn{1}{c|}{M} & \multicolumn{1}{c|}{L} & \multicolumn{1}{c|}{mAP} & \multicolumn{1}{c|}{S} & \multicolumn{1}{c|}{M} & \multicolumn{1}{c|}{L} & \multicolumn{1}{c|}{mAP} & \multicolumn{1}{c|}{S} & \multicolumn{1}{c|}{M} & \multicolumn{1}{c|}{L} & \multicolumn{1}{c|}{mAP} & \multicolumn{1}{c|}{S} & \multicolumn{1}{c|}{M} & \multicolumn{1}{c|}{L} & \multicolumn{1}{c|}{mAP} & \multicolumn{1}{c|}{S} & \multicolumn{1}{c|}{M} & \multicolumn{1}{c|}{L} \\ \hline
		SSD                                     & VGG16                                         & \textbf{79.1}            & -                      & \textbf{35.5}          & 81.9                   & 85.7                     & -                      & 55.5                   & 87.2                   & 87.1                     & \textbf{45.5}          & 61.1                   & 89.6                   & 84.0                     & \textbf{13.1}          & 65.3                   & 89.4                   & 79.0                     & 27.0                   & 68.9                   & \textbf{88.1}          \\
		F-SSD                                   & VGG16                                         & 77.8                     & -                      & 30.2                   & 82.7                   & \textbf{86.1}            & -                      & \textbf{64.6}          & \textbf{88.1}          & 87.9                     & 26.6                   & 56.9                   & \textbf{90.3}          & \textbf{86.1}            & 9.8                    & \textbf{66.9}          & 89.4                   & 78.6                     & 25.6                   & 67.3                   & 88.0                   \\
		A-SSD                                   & VGG16                                         & 78.0                     & -                      & 28.0                   & 81.9                   & 85.5                     & -                      & 64.2                   & 87.5                   & 86.9                     & 28.2                   & 57.3                   & 89.5                   & 84.1                     & 9.2                    & 60.7                   & \textbf{90.0}          & \textbf{79.5}            & \textbf{29.7}          & \textbf{70.7}          & \textbf{88.1}          \\
		FA-SSD                                  & VGG16                                         & 77.8                     & -                      & 25.7                   & \textbf{83.2}          & 85.8                     & -                      & 60.1                   & 87.6                   & \textbf{88.2}            & 26.6                   & \textbf{65.1}          & \textbf{90.3}          & 85.0                     & 13.0                   & 64.4                   & 89.7                   & 79.2                     & 29.1                   & 68.3                   & \textbf{88.1}          \\ \hline
		SSD                                     & Resnet18                                      & 72.1                     & -                      & 18.2                   & 79.3                   & 77.5                     & -                      & 30.8                   & 84.6                   & 80.8                     & 12.1                   & 32.7                   & \textbf{89.8}          & 78.7                     & 0.0                    & 48.3                   & \textbf{89.1}          & 73.0                     & 11.0                   & 55.2                   & 87.2                   \\
		F-SSD                                   & Resnet18                                      & \textbf{75.1}            & -                      & 10.4                   & 79.2                   & \textbf{83.8}            & -                      & 48.7                   & \textbf{87.0}          & 85.2                     & 18.9                   & 39.9                   & \textbf{89.8}          & \textbf{83.1}            & \textbf{18.2}          & 52.5                   & 89.4                   & 74.8                     & 18.2                   & 61.3                   & 87.3                   \\
		A-SSD                                   & Resnet18                                      & 74.2                     & -                      & \textbf{27.0}          & 78.5                   & \textbf{83.8}            & -                      & 50.3                   & 86.5                   & 85.1                     & \textbf{46.4}          & 42.7                   & 89.4                   & 82.7                     & 0.0                    & 56.3                   & 88.9                   & 74.8                     & \textbf{18.6}          & 61.5                   & \textbf{87.7}          \\
		FA-SSD                                  & Resnet18                                      & 73.7                     & -                      & 18.2                   & \textbf{80.5}          & 84.4                     & -                      & \textbf{52.2}          & 86.5                   & \textbf{85.8}            & 25.5                   & \textbf{42.9}          & 89.0                   & 82.8                     & 0.0                    & \textbf{59.6}          & 89.2                   & \textbf{75.1}            & 16.7                   & \textbf{63.6}          & 87.4                   \\ \hline
		SSD                                     & Resnet34                                      & 74.9                     & -                      & 24.3                   & \textbf{81.5}          & 84.0                     & -                      & 49.2                   & 86.2                   & 87.2                     & 44.2                   & 43.3                   & 90.0                   & 79.6                     & 0.0                    & 52.9                   & 89.6                   & 76.0                     & 11.9                   & 61.1                   & \textbf{88.5}          \\
		F-SSD                                   & Resnet34                                      & 75.2                     & -                      & 27.0                   & 79.0                   & \textbf{85.6}            & -                      & 58.0                   & \textbf{87.7}          & 87.8                     & \textbf{18.2}          & 49.2                   & 90.3                   & \textbf{84.8}            & 0.0                    & 60.5                   & \textbf{90.0}          & 76.2                     & \textbf{18.9}          & 65.8                   & 87.8                   \\
		A-SSD                                   & Resnet34                                      & \textbf{76.8}            & -                      & 26.4                   & 81.4                   & 83.6                     & -                      & 61.3                   & 85.8                   & 86.3                     & 15.9                   & 55.0                   & \textbf{90.4}          & 79.9                     & 0.0                    & 55.4                   & 89.8                   & 75.9                     & 18.5                   & 65.1                   & 88.0                   \\
		FA-SSD                                  & Resnet34                                      & 73.2                     & -                      & 20.3                   & 80.0                   & 85.2                     & -                      & \textbf{61.7}          & 87.0                   & \textbf{87.9}            & 12.1                   & \textbf{57.0}          & 89.8                   & 79.9                     & \textbf{27.3}          & \textbf{61.3}          & 89.7                   & \textbf{76.6}            & 17.9                   & \textbf{66.4}          & 88.4                   \\ \hline
		SSD                                     & Resnet50                                      & 74.4                     & -                      & 31.5                   & 81.1                   & 85.2                     & -                      & 54.1                   & 87.7                   & 86.6                     & \textbf{31.8}          & 44.6                   & 89.9                   & 79.9                     & 0.0                    & 52.8                   & 89.9                   & 76.2                     & 12.7                   & 62.6                   & \textbf{88.3}          \\
		F-SSD                                   & Resnet50                                      & 76.7                     & -                      & 27.6                   & 81.7                   & 86.8                     & -                      & \textbf{64.4}          & 88.0                   & \textbf{88.4}            & 0.0                    & 50.7                   & \textbf{90.4}          & \textbf{86.8}            & \textbf{5.5}           & \textbf{64.9}          & \textbf{90.5}          & 77.1                     & 22.5                   & 69.0                   & 88.2                   \\
		A-SSD                                   & Resnet50                                      & 77.2                     & -                      & 28.1                   & 80.9                   & 87.5                     & -                      & 58.2                   & 88.6                   & 87.5                     & 18.2                   & \textbf{54.1}          & \textbf{90.4}          & 85.4                     & 0.0                    & 60.5                   & 90.4                   & \textbf{77.6}            & 24.3                   & 70.2                   & \textbf{88.3}          \\
		FA-SSD                                  & Resnet50                                      & \textbf{79.1}            & -                      & \textbf{37.3}          & \textbf{82.4}          & \textbf{87.6}            & -                      & 62.8                   & \textbf{88.9}          & 88.1                     & 18.2                   & 49.2                   & 90.2                   & 86.3                     & 0.0                    & 62.2                   & 90.3                   & \textbf{77.6}            & \textbf{24.4}          & \textbf{70.6}          & \textbf{88.3}          \\ \hline
		\multicolumn{1}{|c|}{\multirow{2}{*}{}} & \multicolumn{1}{c|}{\multirow{2}{*}{Network}} & \multicolumn{4}{c|}{Potted plant}                                                                   & \multicolumn{4}{c|}{Sheep}                                                                          & \multicolumn{4}{c|}{Sofa}                                                                           & \multicolumn{4}{c|}{Train}                                                                          & \multicolumn{4}{c|}{TV monitor}                                                                     \\ \cline{3-22} 
		\multicolumn{1}{|c|}{}                  & \multicolumn{1}{c|}{}                         & \multicolumn{1}{c|}{mAP} & \multicolumn{1}{c|}{S} & \multicolumn{1}{c|}{M} & \multicolumn{1}{c|}{L} & \multicolumn{1}{c|}{mAP} & \multicolumn{1}{c|}{S} & \multicolumn{1}{c|}{M} & \multicolumn{1}{c|}{L} & \multicolumn{1}{c|}{mAP} & \multicolumn{1}{c|}{S} & \multicolumn{1}{c|}{M} & \multicolumn{1}{c|}{L} & \multicolumn{1}{c|}{mAP} & \multicolumn{1}{c|}{S} & \multicolumn{1}{c|}{M} & \multicolumn{1}{c|}{L} & \multicolumn{1}{c|}{mAP} & \multicolumn{1}{c|}{S} & \multicolumn{1}{c|}{M} & \multicolumn{1}{c|}{L} \\ \hline
		SSD                                     & VGG16                                         & 50.7                     & 6.2                    & 41.2                   & 63.1                   & 77.7                     & 30.3                   & 74.4                   & \textbf{83.6}          & 78.9                     & -                      & \textbf{49.4}          & 81.6                   & 86.2                     & -                      & 63.7                   & 87.8                   & \textbf{76.7}            & 48.6                   & \textbf{67.6}          & 82.6                   \\
		F-SSD                                   & VGG16                                         & 51.0                     & 8.0                    & \textbf{43.5}          & 63.2                   & \textbf{79.0}            & \textbf{49.9}          & \textbf{82.2}          & 81.9                   & \textbf{81.5}            & -                      & 39.4                   & \textbf{82.7}          & \textbf{87.2}            & -                      & \textbf{67.3}          & \textbf{89.4}          & 75.2                     & \textbf{52.6}          & 63.8                   & 83.4                   \\
		A-SSD                                   & VGG16                                         & 50.4                     & \textbf{13.0}          & 41.7                   & 62.3                   & 77.1                     & 35.6                   & 76.7                   & 82.8                   & 79.6                     & -                      & 45.8                   & 80.4                   & 85.8                     & -                      & 64.7                   & 88.0                   & 76.5                     & 36.4                   & 66.9                   & 81.5                   \\
		FA-SSD                                  & VGG16                                         & \textbf{51.1}            & 9.3                    & 41.8                   & \textbf{63.6}          & 76.4                     & 39.4                   & 81.4                   & 78.9                   & 79.9                     & -                      & 40.9                   & 81.9                   & 86.4                     & -                      & 56.8                   & 88.5                   & 76.2                     & 50.0                   & 63.9                   & \textbf{85.1}          \\ \hline
		SSD                                     & Resnet18                                      & 40.0                     & 0.2                    & 28.0                   & 53.6                   & 69.2                     & 25.0                   & 61.1                   & 78.3                   & 74.5                     & -                      & 23.6                   & 79.0                   & 78.0                     & -                      & 48.6                   & 84.1                   & 69.4                     & 27.3                   & 50.5                   & 81.5                   \\
		F-SSD                                   & Resnet18                                      & \textbf{46.0}            & 1.8                    & 38.0                   & \textbf{58.5}          & \textbf{73.4}            & 22.5                   & \textbf{70.5}          & \textbf{79.3}          & \textbf{78.8}            & -                      & 36.6                   & \textbf{80.3}          & \textbf{84.3}            & -                      & 66.0                   & 85.9                   & \textbf{73.6}            & 33.7                   & 61.7                   & 82.6                   \\
		A-SSD                                   & Resnet18                                      & 44.1                     & 1.0                    & 35.2                   & 57.6                   & 71.1                     & 23.9                   & 68.7                   & 77.5                   & 78.2                     & -                      & \textbf{40.5}          & 79.4                   & 83.3                     & -                      & 54.2                   & 86.0                   & 72.6                     & 28.8                   & 58.0                   & \textbf{84.0}          \\
		FA-SSD                                  & Resnet18                                      & \textbf{46.0}            & \textbf{5.3}           & \textbf{38.2}          & \textbf{58.5}          & 73.3                     & \textbf{25.0}          & 70.4                   & 79.1                   & 77.4                     & -                      & 28.7                   & 79.4                   & 84.0                     & -                      & \textbf{70.4}          & \textbf{86.2}          & 72.7                     & \textbf{46.9}          & \textbf{61.9}          & 83.1                   \\ \hline
		SSD                                     & Resnet34                                      & 49.1                     & 2.3                    & 41.2                   & 61.7                   & 74.1                     & 19.7                   & 68.8                   & 85.2                   & 80.2                     & -                      & 36.4                   & 81.4                   & 85.0                     & -                      & 57.8                   & 87.7                   & 70.7                     & 24.2                   & 57.2                   & 79.2                   \\
		F-SSD                                   & Resnet34                                      & \textbf{53.8}            & 2.3                    & \textbf{46.2}          & 66.0                   & \textbf{76.8}            & \textbf{36.6}          & 75.7                   & \textbf{87.9}          & \textbf{80.9}            & -                      & 38.5                   & \textbf{82.7}          & 85.7                     & -                      & 65.1                   & \textbf{88.0}          & \textbf{76.1}            & 33.5                   & 62.9                   & \textbf{84.1}          \\
		A-SSD                                   & Resnet34                                      & 49.5                     & \textbf{4.7}           & 39.3                   & 61.8                   & 74.2                     & 32.2                   & 67.0                   & 85.0                   & 78.8                     & -                      & \textbf{40.6}          & 80.4                   & 84.4                     & -                      & 58.6                   & 87.3                   & 72.6                     & 27.3                   & 64.1                   & 82.0                   \\
		FA-SSD                                  & Resnet34                                      & 53.5                     & 3.3                    & 45.2                   & \textbf{66.3}          & 76.2                     & 27.6                   & \textbf{80.4}          & 85.2                   & 79.4                     & -                      & 28.5                   & 81.0                   & \textbf{87.0}            & -                      & \textbf{65.4}          & 87.9                   & \textbf{76.1}            & \textbf{55.2}          & \textbf{66.7}          & 79.8                   \\ \hline
		SSD                                     & Resnet50                                      & 48.4                     & 2.7                    & 33.3                   & 65.2                   & 75.4                     & 26.9                   & 74.5                   & 85.0                   & 80.8                     & -                      & 36.0                   & \textbf{82.9}          & 86.2                     & -                      & 55.6                   & 87.9                   & 72.6                     & 22.1                   & 57.7                   & 79.9                   \\
		F-SSD                                   & Resnet50                                      & 51.4                     & 7.3                    & 46.7                   & 60.0                   & 78.7                     & 29.2                   & \textbf{83.7}          & \textbf{87.4}          & 79.9                     & -                      & 27.7                   & 81.0                   & 86.9                     & -                      & \textbf{61.6}          & \textbf{88.5}          & 77.0                     & 43.5                   & \textbf{71.0}          & \textbf{87.0}          \\
		A-SSD                                   & Resnet50                                      & \textbf{54.4}            & 1.7                    & 46.9                   & \textbf{66.5}          & \textbf{81.0}            & \textbf{42.5}          & 77.6                   & 87.0                   & 80.7                     & -                      & 39.4                   & \textbf{82.9}          & \textbf{87.2}            & -                      & 77.3                   & 88.1                   & 75.9                     & 44.6                   & 67.2                   & 84.0                   \\
		FA-SSD                                  & Resnet50                                      & 53.3                     & \textbf{9.0}           & \textbf{47.7}          & 64.2                   & 78.5                     & 30.9                   & 82.2                   & 87.2                   & \textbf{81.5}            & -                      & \textbf{42.6}          & \textbf{82.9}          & 86.2                     & -                      & 60.2                   & 87.9                   & \textbf{77.4}            & \textbf{49.6}          & 70.8                   & 86.0                   \\ \hline
	\end{tabular}
\end{table*}

\end{document}